\pdfoutput=1

\documentclass[11pt]{article}

\usepackage[final]{emnlp2021}

\usepackage{times}
\usepackage{latexsym}

\usepackage[T1]{fontenc}

\usepackage[utf8]{inputenc}

\usepackage{microtype}

\usepackage[ruled,vlined]{algorithm2e}
\usepackage{times} 
\usepackage{latexsym}
\usepackage{amssymb}		%
\usepackage{graphicx}		%
\usepackage{color}
\usepackage{natbib}
\usepackage{bookmark}
\usepackage{natbib}
\usepackage{eucal}
\usepackage{amsmath}
\usepackage{multirow}
\usepackage{booktabs}
\usepackage{subcaption}
\usepackage{enumitem}
\usepackage{xspace}
\usepackage{ragged2e}
\usepackage{array}
\usepackage{amsthm}
\usepackage{algorithmic}
\usepackage{soul}

\newif\ifhidecomments
\hidecommentsfalse

\theoremstyle{definition}
\newtheorem{definition}{Definition}

\ifhidecomments
    \newcommand{\joe}[1]{}
    \newcommand{\chenhao}[1]{}
\else
    \newcommand{\joe}[1]{{\color{red}{ #1 --Joe}}}
    \newcommand{\chenhao}[1]{{\color{blue}{\tt #1 --CT}}}
\fi

\newcommand{\para}[1]{\smallskip\noindent{\bf #1}}
\newcommand{\secref}[1]{\S\ref{#1}}
\newcommand{\figref}[1]{Fig.~\ref{#1}}
\newcommand{\tableref}[1]{Table~\ref{#1}}
\newcommand{\system}{\texttt{DecSum}\xspace}
\newcommand{\msewithfull}{\textit{MSE with full}\xspace}
\newcommand{\meaningdiversity}{textual non-redundancy\xspace}
\newcommand{\summarynotation}{\ensuremath{\tilde{X}}}
\newcommand{\trainingset}{\ensuremath{D_{\operatorname{train}}}}

\title{Decision-Focused Summarization}

\author{Chao-Chun Hsu \\ 
  University of Chicago \\
\texttt{chaochunh@uchicago.edu} \\
\And
  Chenhao Tan \\
  University of Chicago \\
  \texttt{chenhao@uchicago.edu} \\
 }

\begin{document}
\maketitle
\begin{abstract}
Relevance in summarization is typically defined based on textual information alone, {\em without} incorporating insights about a particular decision.
As a result, to support risk analysis of pancreatic cancer, summaries of medical notes may include irrelevant information such as a knee injury.
We propose a novel problem, decision-focused summarization, where the goal is to summarize relevant information {\em for a decision}.
We leverage a predictive model that makes the decision based on the full text to provide valuable insights on how a decision can be inferred from text.
To build a summary, we then select {\em representative} sentences that lead to similar model decisions as using the full text while accounting for \meaningdiversity.
To evaluate our method (\system), 
we build a testbed where the task is to summarize the first ten reviews of a restaurant in support of predicting its future rating on Yelp.
\system substantially outperforms text-only summarization methods and model-based explanation methods in decision faithfulness and representativeness.
We further demonstrate that
\system is the only method that enables humans to outperform random chance in 
predicting which restaurant will be better rated
in the future.

\end{abstract}

\section{Introduction}
\label{sec:intro}

\begin{figure}[t]
    \centering
    \includegraphics[width=0.9\columnwidth]{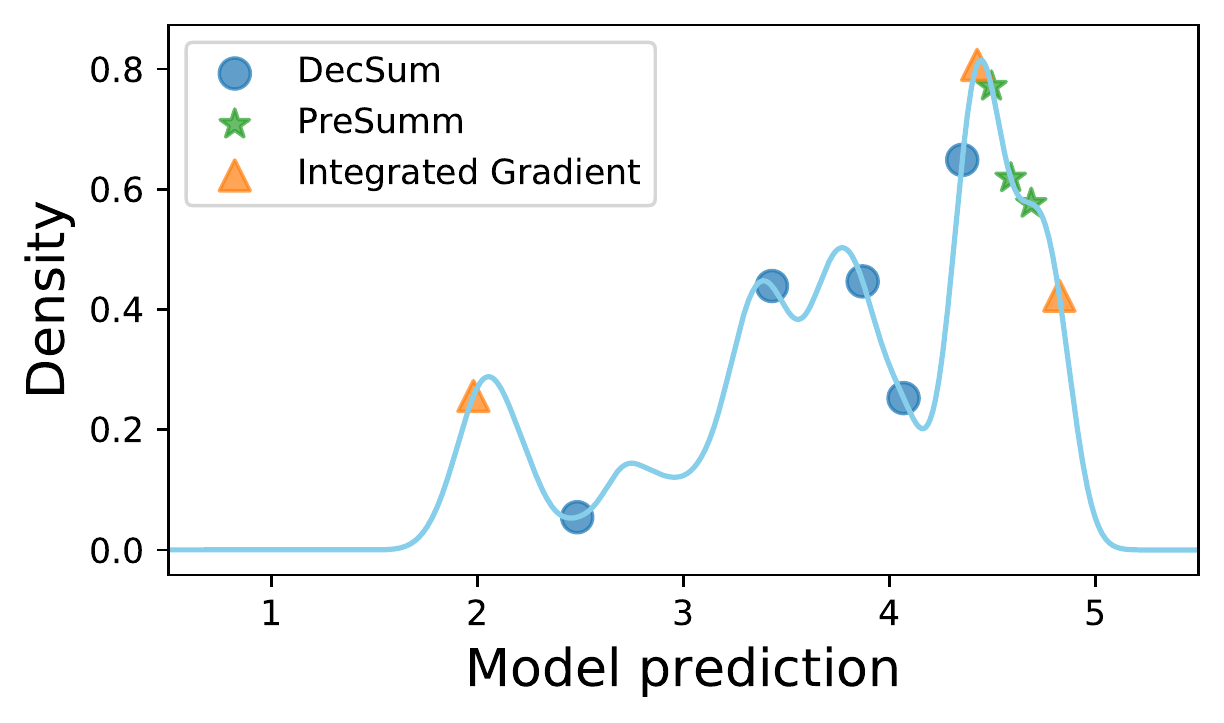} 
    \caption{
    Illustration of the selected sentences by different methods on the distribution of model predictions on all individual sentences.
    Our method (\system) 
    covers 
    the full distribution, while 
    PreSumm, a text-only summarization method, concentrates on the right side, and integrated gradients, a model-explanation method, misses the middle part.
    }
    \label{fig:dist}
\end{figure}

Human decision making often requires making sense of a large amount of information.
For instance, doctors go through a myriad of medical notes to determine the risk of pancreatic cancer, and investors need to decide whether a stock price will increase based on hundreds of analyst reports.
In these cases, summarization can potentially support human decision making by identifying the most relevant information for these decisions \citep{demner2009can,workman2012text}.

Ideally, decision-focused summarization should incorporate insights about how decisions can be inferred from text.
However, typical summarization methods in NLP define relevance based on the textual information exclusively.
An example desideratum is \meaningdiversity \citep{carbonell1998use}, which encourages the summaries to cover diverse information in the input documents.
Fully optimizing this text-only criterion can be counter-productive for decision making:
information about a knee injury does not really help understand the risk of pancreatic cancer,
and the disclaimers in financial analysts may not be the most relevant for investment decisions.

In this work, we investigate the potential of leveraging a supervised decision model for {\em extractive} decision-focused summarization.
A predictive model that learns to make a decision given the full text can encode valuable insights about how the decision can be inferred from text.
Given that \citet{kleinberg2015prediction} shows that many policy problems depend on predictive inference,
incorporating model-based insights into summarization
can be widely applicable to many decisions in high-stake scenarios such as finance and healthcare.

We propose novel desiderata for decision-focused summarization in addition to \meaningdiversity and formalize them based on model behavior.
First, \textit{decision faithfulness} suggests that the selected sentences should lead to the same decision as using the full text based on the model.
This desideratum is analogous to sufficiency in evaluating the interpretability of attribution methods \citep{deyoung2019eraser}, as attribution methods should ideally identify sentences that would ``explain'' the model's decision with all sentences. 
This observation also highlights the connection between explanation and decision-focused summarization.

In addition to faithfulness, {\em decision representativeness} resembles \meaningdiversity in the decision space. 
\figref{fig:dist} illustrates the decision distribution of all individual sentences in the input documents, i.e., model predictions given each sentence, and sentences chosen by different methods.
Ideally, the selected sentences should be representative of this overall decision distribution.
Our method is designed to optimize this desideratum, whereas text-only summarization methods and model-based explanation methods do not aim to select sentences that represent the whole distribution.

To evaluate our proposed method, we formulate a future rating prediction task on Yelp, inspired by investment decisions.
The task is to predict a restaurant's future rating given the first ten reviews.
Automatic metrics demonstrate that our method (\system) outperforms text-only summarization methods and model-based explanation methods in 
decision faithfulness and decision representativeness.
\system also improves \meaningdiversity over the baselines, although at the cost of grammaticality and coherence.
Human evaluation further shows that \system is the {\em only} method that enables humans to statistically outperform random chance in predicting which restaurant will be rated better in the future. 

To summarize, our main contributions are:
\begin{itemize}[itemsep=-2pt]
    \item We propose a novel summarization task that emphasizes supporting decision making.
    \item We propose {\em decision faithfulness} and {\em decision representativeness} as important desiderata for this task in addition to \meaningdiversity, based on the behavior of a supervised model.
    \item Using Yelp future rating prediction as a testbed, we show that the proposed approach outperforms text-only summarization methods and model-based explanation methods.
    \item We show that the proposed approach effectively supports human decision making in a very challenging classification task.
\end{itemize}
\section{Method}
\label{sec:method}

In this section, we formalize decision-focused summarization and three desiderata. 
We then provide a greedy algorithm to optimize the three desiderata.

\subsection{Problem Formulation}

Decision-focused summarization is conditioned on a decision of interest, e.g., whether a stock price will increase.
We refer to this decision as $y$.
It is challenging for humans to make decisions based on the full input text, $X$, which can be hundreds of analyst reports. 
The task is thus to identify the most relevant information from the input for a particular decision as a summary in support of human decision making.
We formulate the extractive version of decision-focused summarization as follows.
\begin{definition}[Decision-focused summarization]
Given an input text $X=\{{x_s}\}_{s=1}^{s=S}$, where $S$ is the number of sentences, select
a subset of sentences $\summarynotation \subset X$ to support making the decision $y$. 
\end{definition}

Unlike typical summarization where we only have access to textual information,
decision-focused summarization requires knowledge of how the decision can be inferred from the text. 
Our problem setup thus has a training set analogous to supervised learning, $\trainingset = \{(X_i, y_i)\}$, which can provide insights on 
the relation between the text and the decision.

\para{Yelp future rating prediction task.} 
Inspired by investment decisions given analyst reports, we consider a future rating prediction task in the context of Yelp as a testbed.
This allows us to have access to both a dataset\footnote{\url{https://www.yelp.com/dataset}.} and participants who may be able to perform this task.
Specifically, for each restaurant in Yelp, we define $X$ as the text of the first $k$ reviews and $y$ is the average rating of the first $t$ reviews where $t > k$ so that the task is to forecast future ratings. 
We use $k=10$ and $t=50$ in this work.
Our problem is then to select sentences from a restaurant's first 10 reviews in support of predicting its future rating after 50 reviews.

\subsection{DecSum}

The key intuition of our approach (\system) is to develop a model that makes the 
decision given the text ($f: X \rightarrow y$) and then build summaries that can both support this model in making accurate decisions and account for properties in text-only summarization.
This model can be seen as a virtual decision maker and hopefully encodes valuable information of how the decision can be inferred from the text.
We obtain $f$ from $\trainingset$ using standard supervised models. 

As discussed in \secref{sec:intro}, decision-focused summaries should satisfy decision faithfulness, decision representativeness, and \meaningdiversity.
Next, we formally define these desiderata as objective (loss) functions that can be minimized to extract decision-focused summaries.

\para{Decision faithfulness.}
The first desideratum is that the selected sentences should lead to similar decisions as the full text: $f(\summarynotation) \simeq f(X)$.
A natural loss function is the absolute difference between $f(\summarynotation)$ and $f(X)$, and here we use its logarithm:
\[\mathcal{L}_{\operatorname{F}}(\summarynotation, X, f) = \log |f(\summarynotation) - f(X)|.
\]
This desideratum resonates with faithfulness in interpretability \citep{jacovi-goldberg-2020-towards}.
However, our focus is not on whether the model {\em actually} uses these sentences in its prediction, but on the behavioral outcome of the sentences, i.e., whether they supports model/human decision making by identifying relevant information for the decision. 

\para{Decision representativeness.}
Sentences in the full input $X$ can lead to very different decisions on their own.
Thus, in addition to 
decision faithfulness, 
model decisions of selected sentences should be representative of the decision distribution of sentences in the full input (\figref{fig:dist}). 
In other words, the decision distribution of the summary $\hat{Y}_{\summarynotation}= \{f(x) \mid x \in \summarynotation\}$ should be close to the decision distribution of all sentences in 
the full text
$\hat{Y}_{X} = \{f(x) \mid x \in X\}$.
To measure the distance between 
$\hat{Y}_{\summarynotation}$ and $\hat{Y}_{X}$,
we use 
the Wasserstein Distance~\cite{Ramdas_2017}:%
\[\scriptstyle W(\hat{Y}_{\summarynotation},\hat{Y}_{X}) = \mathop{\text{inf}}_{\gamma \in \Gamma(\hat{Y}_{\summarynotation},\hat{Y}_{X})} \int_{\mathbb{R} \times \mathbb{R}} ||f - f'|| d\gamma(f, f'),
\]
where $\Gamma (\hat{Y}_{\summarynotation},\hat{Y}_{X})$ denotes the collection of all measures on $ \mathbb{R}\times \mathbb{R}$ with marginals $\hat{Y}_{\summarynotation}$ and $\hat{Y}_{X}$  on the first and second factors respectively.
Our second loss function is then the logarithm of the Wasserstein distance between the decision distribution of the summary and that of the full text:
\[
    \mathcal{L}_{\operatorname{R}}(\summarynotation, X, f) = \log(W(\hat{Y}_{\summarynotation},\hat{Y}_{X})).
\]

\para{Textual non-redundancy.}
Our third desired property is inspired by prior work on diversity in textual summarization: the selected sentences should capture diverse contents and provide an overview of the textual information in the input text \citep{lin2011class,dasgupta2013summarization,carbonell1998use}.
To operationalize this intuition, we adopt a loss function to
encourage sentences in the summary to be dissimilar to each other.
We opertationalize similarity using the cosine similarity based on SentBERT  sentence representation $s(x)$ \cite{reimers-2019-sentence-bert}: 
\[
    \mathcal{L}_{\operatorname{D}}(\summarynotation) = \sum_{x \in \summarynotation} \max_{x' \in \summarynotation - \{x\}}  \mathrm{cossim}(s(x), s(x')).
\]
To summarize, our objective function consists of the above three parts:
\[
        \scriptstyle
    \mathcal{L}(\summarynotation, X, f) = \alpha \mathcal{L}_{\operatorname{F}}(\summarynotation, X, f) + \beta  \mathcal{L}_{\operatorname{R}}(\summarynotation, X, f) + \gamma \mathcal{L}_{\operatorname{D}}(\summarynotation),
\]
\noindent where $\alpha, \beta, \gamma$ control the tradeoff between the three desiderata.
Note that decision faithfulness ($\mathcal{L}_{\operatorname{F}}$) and decision representativeness ($\mathcal{L}_{\operatorname{R}}$) both rely on $f$, while \meaningdiversity ($\mathcal{L}_{\operatorname{D}}$) depends on the textual information alone.
We use $\log$ in $\mathcal{L}_{\operatorname{F}}$ and $\mathcal{L}_{\operatorname{R}}$ because they are unbounded.

\para{Algorithm implementation.}
Inspired by traditional summarization methods \cite{carbonell1998use,mihalcea-tarau-2004-textrank}, we develop an iterative algorithm that greedily selects a sentence that 
minimizes our loss function.
A key advantage of this approach is that it exposes the design space and presents a white box for researchers.

Algorithm~\ref{algo:sent-select} shows the full algorithm.
To select $K$ sentences from input $X$, in each step $k=\{1,...,K\}$, we iteratively choose a sentence among the remaining sentences, $\hat{x} \in X - \summarynotation_{k-1}$, that achieves the lowest loss $\mathcal{L}(\summarynotation_{k-1} \cup \{\hat{x}\}, X, f)$ where $\summarynotation_{k-1}$ is the current summary with $k-1$ sentences. 
When $\beta > 0$, we only use 
$\mathcal{L}_R$ at the first step to encourage the algorithm to explore the full distribution rather than stalling at the sentence that is most faithful to $f(X)$.
In practice, we use beam search with beam size of 4 to improve our greedy algorithm.
Our code and data are available at \url{https://github.com/ChicagoHAI/decsum}.

\begin{algorithm}[t]
    \small
    \SetAlgoLined
    \KwIn{$X, f, K$ }
    \KwOut{$\summarynotation$} 
    $\summarynotation  \gets \emptyset$,~~$k  \gets 1$\;
    \While{$k \leq K$}{
        \eIf {$\beta > 0$ \text{ and } $k = 1$}{
            $\hat{x}\gets\text{argmin}_{\hat{x} \in X} \mathcal{L}_{\operatorname{R}}(\{\hat{x}\}, X, f)$
        }{
            $\hat{x} \gets \text{argmin}_{\hat{x} \in X} \mathcal{L}(\summarynotation \cup \{\hat{x}\}, X, f)$
        }
        $\summarynotation \gets \summarynotation \cup \{\hat{x}\}$\;
        $X \gets X - \{\hat{x}\}$\;
        $k \gets k+1$
    }
    \caption{\system}
    \label{algo:sent-select}
\end{algorithm}

\section{Experiment Setup}

Our approach is contingent on a machine learning model that can make decisions based on the input text.
In this section, we discuss our dataset split and choice of this ML model, baselines summarization approaches, and evaluation strategies.

\subsection{Regression Model and Baselines}

We split the Yelp dataset (18,112 restaurants) into 
training/validation/test sets with 64\%/16\%/20\% ratio.
Since the text of 10 reviews has 1,621 tokens on average, we use Longformer \cite{beltagy2020longformer} to fine-tune a regression model.
See details of hyperparameter tuning in the appendix.

In addition to Longformer, we also considered logistic regression and deep averaging networks \citep{Iyye2015rdan} for this problem.
However, we find that only Longformer leads to an appropriate distribution of the predicted score ($f(x)$) at the sentence level (see the appendix), suggesting that Longformer may better generalize to shorter inputs.
We refer to this model as the regression model or $f$ to differentiate from summarization methods.

We consider two types of baselines: text-only summarization 
and model-based explanation. 

\para{Text-only summarization baselines.}
We compare \system with both extractive and abstractive summarization methods.

\begin{itemize}[leftmargin=*,itemsep=-2pt,topsep=0pt]
    \item \textbf{PreSumm} is an extractive 
    summarization method with hierarchical encoders
    \cite{Liu_2019}. We use {distilbert-base-uncased}\footnote{\url{https://transformersum.readthedocs.io/en/latest/extractive/models-results.html}.}  
    built on the CNN/DM dataset \cite{hermann2015teaching}, as DistillBERT 
    is competitive with
    BERT.
    \item \textbf{BART} is a seq2seq model trained with a denoising objective \cite{Lewis_2020}. We use {bart-large-cnn} model fine-tuned on CNN/DM. %
    \item \textbf{Random} simply selects random sentences from the input reviews. This method can extract somewhat representative sentences, and we hypothesize that it may be competitive against PreSumm and BART in this task. 
\end{itemize}

\para{Model-based explanations.}
PreSumm and BART do not depend on our regression model, we 
thus consider
attribution methods based on the same model that \system uses as the second type of baselines. 
These attribution methods used in \citet{jain-etal-2020-learning} are supposed to extract sentences that explain the model decision. 

\begin{itemize}[leftmargin=*,itemsep=-2pt,topsep=0pt]
    \item \textbf{Integrated Gradients (IG)} is 
    a gradient-based method \cite{sundararajan2017axiomatic}.
    Following \citet{jain-etal-2020-learning}, we sum up the importance score of input tokens for each sentence and select top $K$ sentences as the results of IG. 
    \item \textbf{Attention} 
    may also be used to interpret transformers.
    We use the mean attention weights of all 12 heads for the \texttt{[CLS]} token at the last layer in Longformer as importance scores for each token, following \citet{jain2020learning}. 
    Similar to IG, we rank sentences based on the summed importance scores over tokens in a sentence.
\end{itemize}

\system, PreSumm, IG, Attention, and Random can all generate a ranking/order for sentences and allow us to control the summary length.

\subsection{Evaluation Metrics and Setup}
\label{sec:evaluation}

Our evaluation consists of both automatic metrics and human evaluations.
All the evaluations are based on the test set, similar to supervised learning.

\para{Automatic metrics.}
We design evaluation metrics based on our three desiderata.

\begin{itemize}[leftmargin=*,itemsep=-2pt,topsep=0pt]
    \item \textbf{Faithfulness to the original model prediction.}
    We rely on the regression model trained based on the full text of the first 10 reviews to measure faithfulness.
    Specifically, we measure the mean squared error between the predicted score based on the summary with the predicted score of the full text, $(f(\summarynotation) - f(X))^2$.
    \item \textbf{Representativeness compared to the decision distribution of all sentences.}
    We measure the Wasserstein distance between the distribution of model predictions of the summary $\hat{Y}_{\summarynotation}$ and 
    that of all sentences in the first 10 reviews $\hat{Y}_{X}$.
    \item \textbf{Text-only summary evaluation metrics.} We use SUM-QE~\cite{Xenouleas:EMNLP-IJCNLP19}, 
    BERT-based 
    automatic summarization evaluation,
    to evaluate five aspects, i.e., grammaticality, non-redundancy, referential clarity, focus, and structure \& coherence. 
    Note that coherence of decision-focused summaries may differ from that of typical summaries, as they are supposed to provide diverse and even conflicting opinions.
\end{itemize}

In addition, we also use MSE with the restaurant rating after 50 reviews to measure the quality of the summaries in the forecasting task, $(f(\summarynotation) - y)^2$.

\begin{figure*}[!t]
    \centering
    \includegraphics[width=0.7\textwidth]{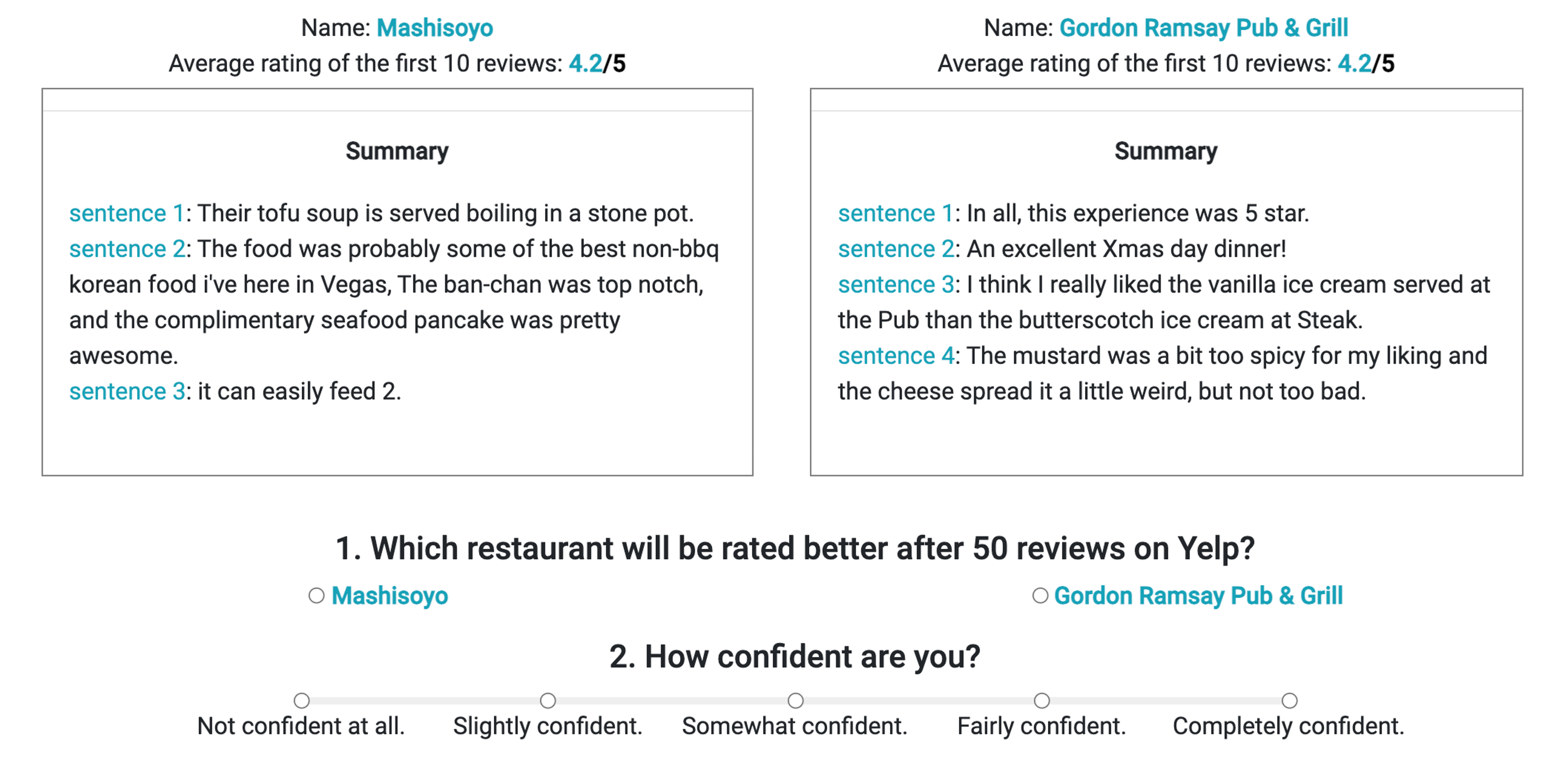}
    \caption{Screenshot of the experiment interface for human evaluation.
    Participants are asked to predict which restaurant will be rated higher after 50 reviews based on the summaries of the first 10 reviews where these two restaurants have the same average rating in the first 10 reviews.
    }
    \label{fig:screenshot}
    \vspace{-8pt}
\end{figure*}

\para{Human evaluation.}
While an obvious idea is asking humans to forecast a restaurant's future rating, this regression task is too challenging for humans.
It is not humans' strength to 
tell the difference between 4.1 and 4.2 in average restaurant ratings.
Therefore, inspired by prior work on pairwise tasks \citep{tan+etal:16a,tan+lee+pang:14,zhang-etal-2018-conversations}, we develop a simplified pairwise classification task: 
given a pair of restaurants with the same average rating of first 10 reviews, we ask participants to guess which will be rated better after 50 reviews.
We ensure 
that these two restaurants are located in the same city and their rating difference is at least one star after 50 reviews. %
1,028 restaurant pairs from the test set
satisfy these criteria, and we randomly select 200 pairs for our human evaluation and limit the number of pairs per city to 25.

We use Mechanical Turk to conduct our human evaluation.
A crowdworker is shown task instructions, an example pair, 10 pairs of restaurants (main task), and an exit survey.
\figref{fig:screenshot} illustrates the experiment interface of the main task.
We only allow participants who have 99\% or more HITs acceptance rate, have 50 or more finished HITs, and are located in the US.
We also require turkers to spend at least 20 seconds for each pair (the hourly salary is $\sim$\$10).
Participants enjoyed our tasks and reported their heuristics in decision making.
See appendix for more details of our experiments.
We collect three human guesses for each pair and consider four summarization methods.
In addition to random\footnote{We considered using the full text of 10 reviews as a baseline. However, participants in pilot studies found the information too overwhelming. Summaries consisting of random sentences provide a more comparable baseline as \system.} and \system,
we choose one text-only summarization method (PreSumm) and one model-based explanation method (IG) according to automatic metrics (see \secref{sec:automatic}).

To make sure that the summaries of different methods are comparable to each other, we control for token length in summaries.
Recall that the summarization length of BART model is not 
easily 
controllable.
Thus, we constrain token length of summaries to the average of BART summaries. %
Specifically, we sequentially select sentences until their length exceeds 50 tokens in the other methods. 
For \system, we set $K=15$ in beam search and then truncate the same way as other methods.
As a result, the summaries from all methods are comparable in length (see the appendix).

\section{Results}

In this section, we compare the quality of summaries from our proposed decision-focused summarization with other existing approaches, both through automatic evaluation metrics and human evaluation.
Automatic metrics show that \system provides better decision faithfulness, decision representativeness, \meaningdiversity than other baselines, but sacrifices other text-only qualities such as coherence and grammaticality.
Human evaluation shows that \system also leads to better human decision making.

\subsection{Automatic Evaluation}
\label{sec:automatic}

We next evaluate three desired properties in \secref{sec:evaluation}.

\subsubsection{Decision Faithfulness}

We 
measure
faithfulness 
by comparing the prediction derived from the summary with the prediction derived from the 10 reviews (\msewithfull).
\tableref{tab:mse} shows that \system with all components on, ``(1, 1, 1)'',  achieves much better faithfulness than any of other baselines, close to 0.
All the text-only summarization methods have an \msewithfull of about 0.34, more than 100 times as much as that of \system.
Model-based explanation methods, surprisingly, lead to even poorer faithfulness 
than text-only methods (IG: $\sim$0.44; attention: $\sim$0.54).

\para{Effect of different components.} 
Our first component, decision faithfulness, is critical for achieving low \msewithfull (all the underlined numbers are below 0.05).
Furthermore, \meaningdiversity improves \msewithfull over optimizing decision faithfulness alone, suggesting that text-only desiderata can in fact support decision making, at least for the AI decision maker. 

Using only \meaningdiversity (0, 0, 1),
a deep version of Maximum Marginal Relevance~\cite{carbonell1998use}, is not better than other text-only summarization methods, i.e., BART and PreSumm.
Interestingly, 
decision representativeness alone (0, 1, 0) leads to better faithfulness than any other baselines, although not as good as directly optimizing \msewithfull.
Henceforth, we use \system to refer to the system with all components on (1, 1, 1) unless otherwise specified.

\begin{table}[t]
    \centering
    \small
    \begin{tabular}{lrr}
        \toprule
        Method &  \shortstack{MSE with \textbf{Full} \\ (faithfulness) $\downarrow$}  &      MSE $\downarrow$ \\
        \midrule
        Full (oracle) & 0 & 0.135 \\\midrule
        \multicolumn{3}{c}{Text-only summarization methods} \\
        Random  & 0.356& 0.475 \\
        BART & 0.368 & 0.502 \\
        PreSumm & 0.339 & 0.478 \\ \midrule
        \multicolumn{3}{c}{Model-based explanation methods} \\
            IG  & 0.436 & 0.565\\
        Attention  & 0.539 & 0.715\\
        \midrule
        \multicolumn{3}{p{0.9\linewidth}}{\system
        ~ w/ ($\alpha$ decision faithfulness, $\beta$ decision representativeness, $\gamma$ \meaningdiversity)} \\ %
        (1, 1, 1)  & \underline{0.0005}& 0.136 \\
        (1, 1, 0)  & \underline{0.0378}& 0.164 \\ %
        (1, 0, 1) & \textbf{\underline{0.0002}} & 0.135  \\ %
        (0, 1, 1)  & 0.162 & 0.283\\ %
        (1, 0, 0) & \underline{0.0264} & 0.155 \\ 
        (0, 1, 0)  & 0.175 & 0.287\\ %
        (0, 0, 1)  & 0.504 & 0.565\\ %
        \bottomrule
    \end{tabular}
    \caption{%
    MSE of model predictions based on summaries of different methods. 
    \textbf{Full} denotes using all reviews without summarization. 
    }
    \label{tab:mse}
\end{table}

\para{Prediction performance.} We also present the MSE with the ground truth rating after 50 reviews.
As expected, using the full text of all ten reviews 
achieves the best 
MSE compared to summarization methods.
The prediction performance of summaries is aligned with \msewithfull. 
\system leads to the best performance compared to baseline models.
Text-only summarization (PreSumm and BART) provides similar performance as random, and outperforms explanation methods (IG and attention), which again highlights that explanation methods do not lead to good summaries even for model decision making.

\subsubsection{Decision Representativeness}

\begin{figure}[t] 
    \centering
    \includegraphics[width=\columnwidth]{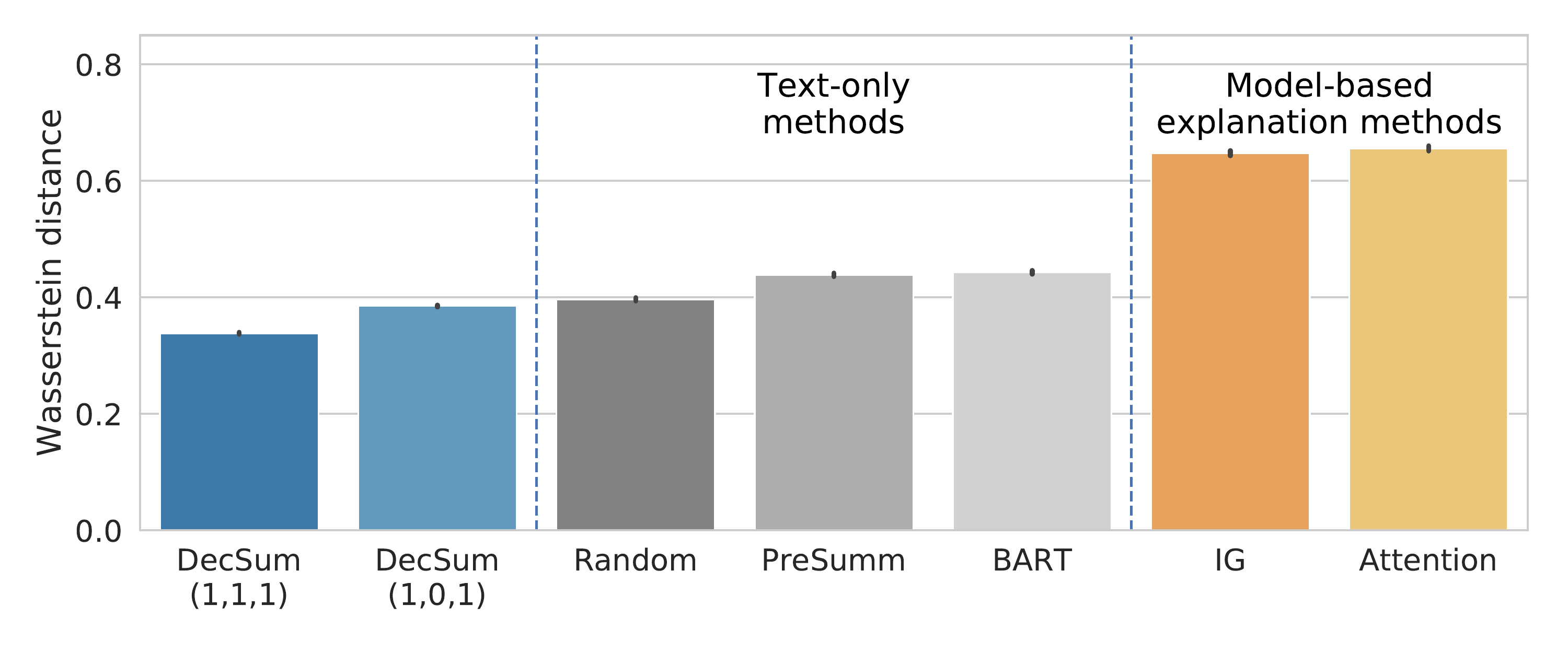}
    \caption{Wasserstien distance between model predictions of summary sentences and all sentences of the first ten reviews. 
    {\em Lower} values indicate better representativeness.
    Error bars represent standard errors.
    \system $(1,1,1)$ is significantly better than other approaches, including \system $(1,0,1)$, with \textit{p}-value $\leq 0.0001$ with paired t-tests.
    }
    \label{fig:wd}
\end{figure}

\begin{figure*}[!t]
    \centering
    \begin{subfigure}[t]{0.32\textwidth}
        \centering
        \includegraphics[width=0.8\textwidth]{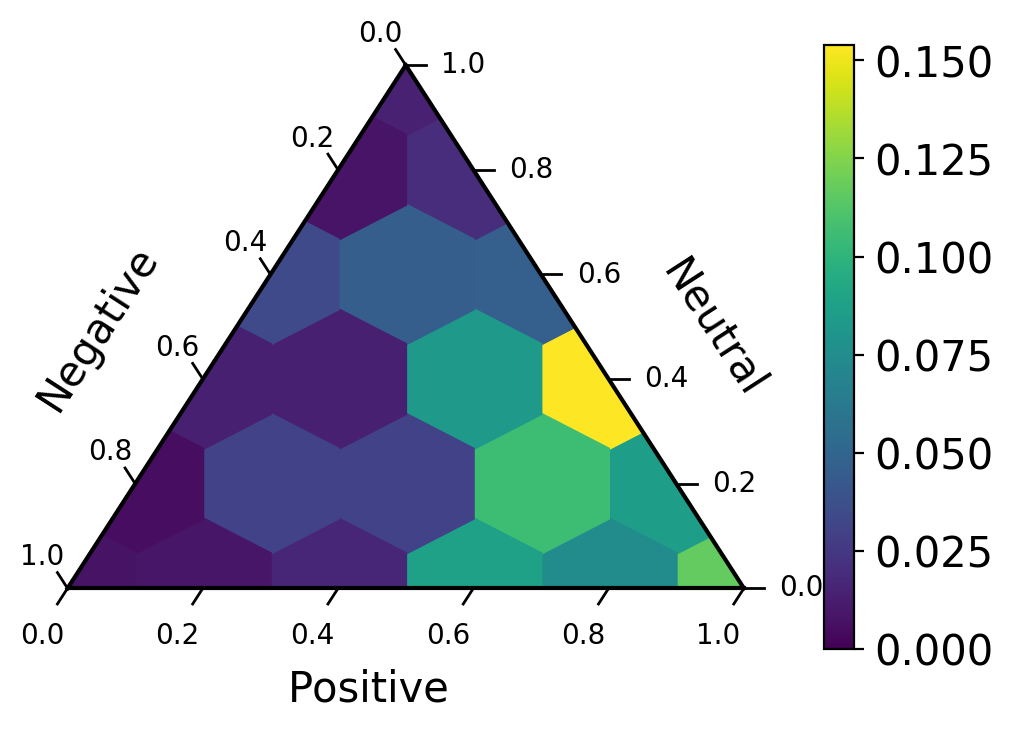}
        \caption{\system (1,1,1).}
        \label{fig:sentiment_decsum}
    \end{subfigure}%
    \begin{subfigure}[t]{0.32\textwidth}
        \centering
        \includegraphics[width=0.8\textwidth]{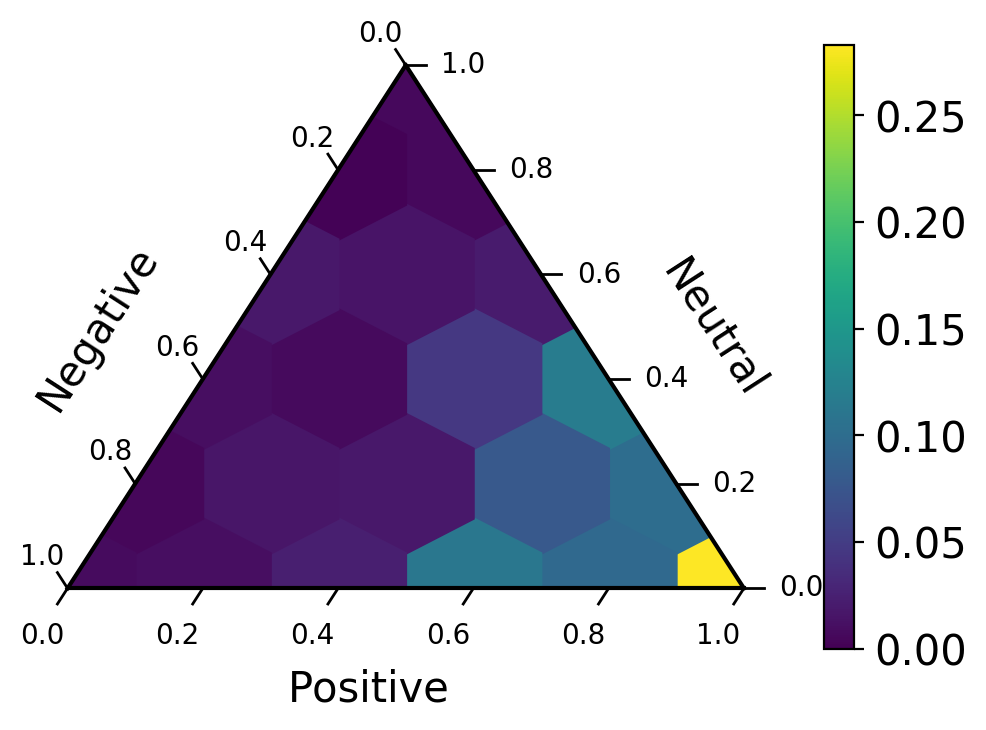}
        \caption{PreSumm.}
        \label{fig:sentiment_presum}
    \end{subfigure}%
    \begin{subfigure}[t]{0.32\textwidth}
        \centering
        \includegraphics[width=0.8\textwidth]{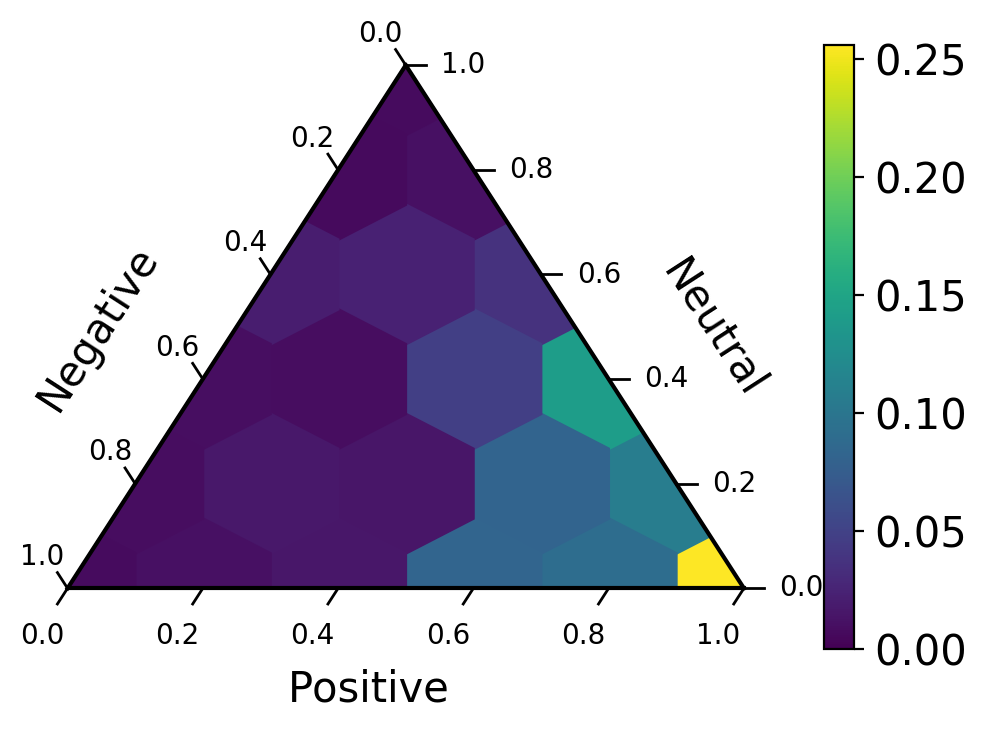}
        \caption{BART.}
        \label{fig:sentiment_bart}
    \end{subfigure}%

    \begin{subfigure}[t]{0.32\textwidth}
        \centering
        \includegraphics[width=0.8\textwidth]{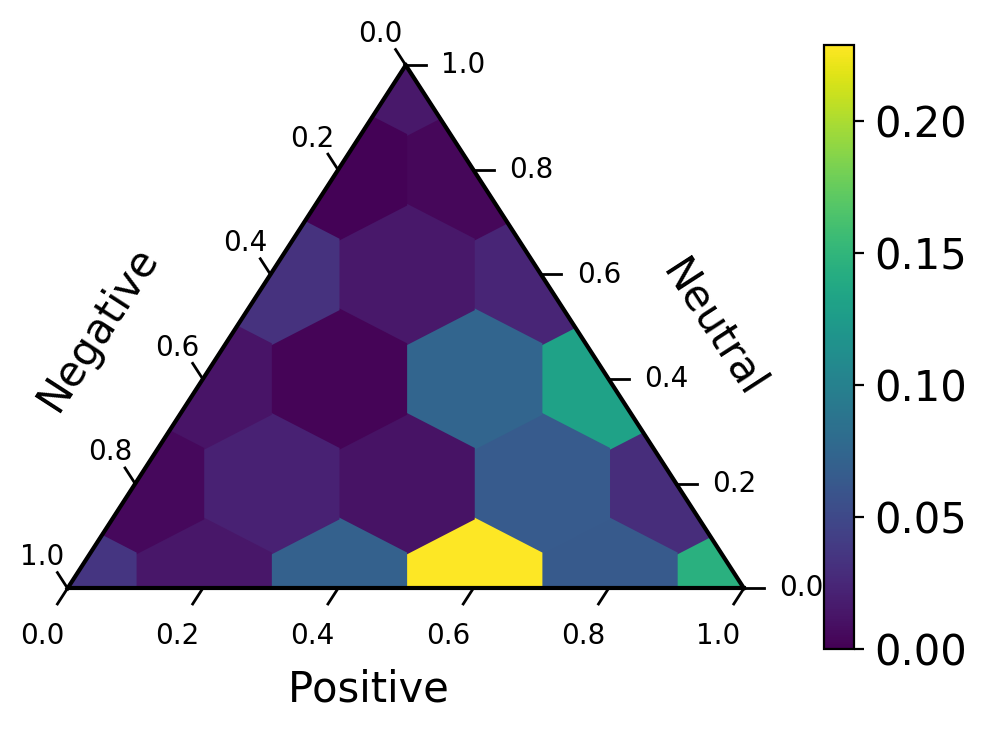}
        \caption{Integrated gradient.}
        \label{fig:sentiment_ig_new}
    \end{subfigure}%
    \begin{subfigure}[t]{0.32\textwidth}
        \centering
        \includegraphics[width=0.8\textwidth]{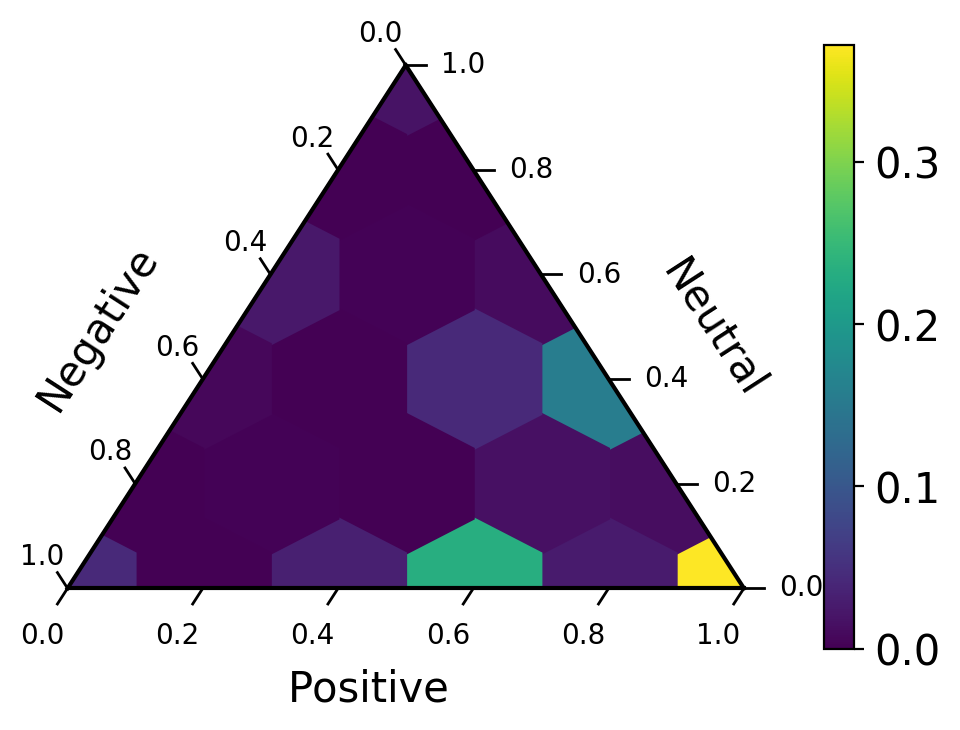}
        \caption{Attention.}
        \label{fig:sentiment_att_new}
    \end{subfigure}%
    \caption{Sentence-level sentiment distribution of summaries. \system can select a wider range of sentences w.r.t. sentiment diversity. 
    }
    \label{fig:sentiment}
\end{figure*}

We start by measuring the Wasserstein distance between model predictions of the selected sentences with those of all the sentences. 
\figref{fig:wd} shows that \system is significantly better than random, text-only summarization, and model-based explanation. 
In other words, \system can select sentences that are more representative of the decision distribution derived from individual sentences in the first ten reviews.
We also compare $(1,1,1)$ with $(1,0,1)$ to examine the effect of the decision representativeness component.
While optimizing decision faithfulness naturally encourages selecting sentences that overall reflect the final decision, the second component further improves the representativeness. 

\begin{figure}[t]
    \centering
    \includegraphics[width=1\columnwidth]{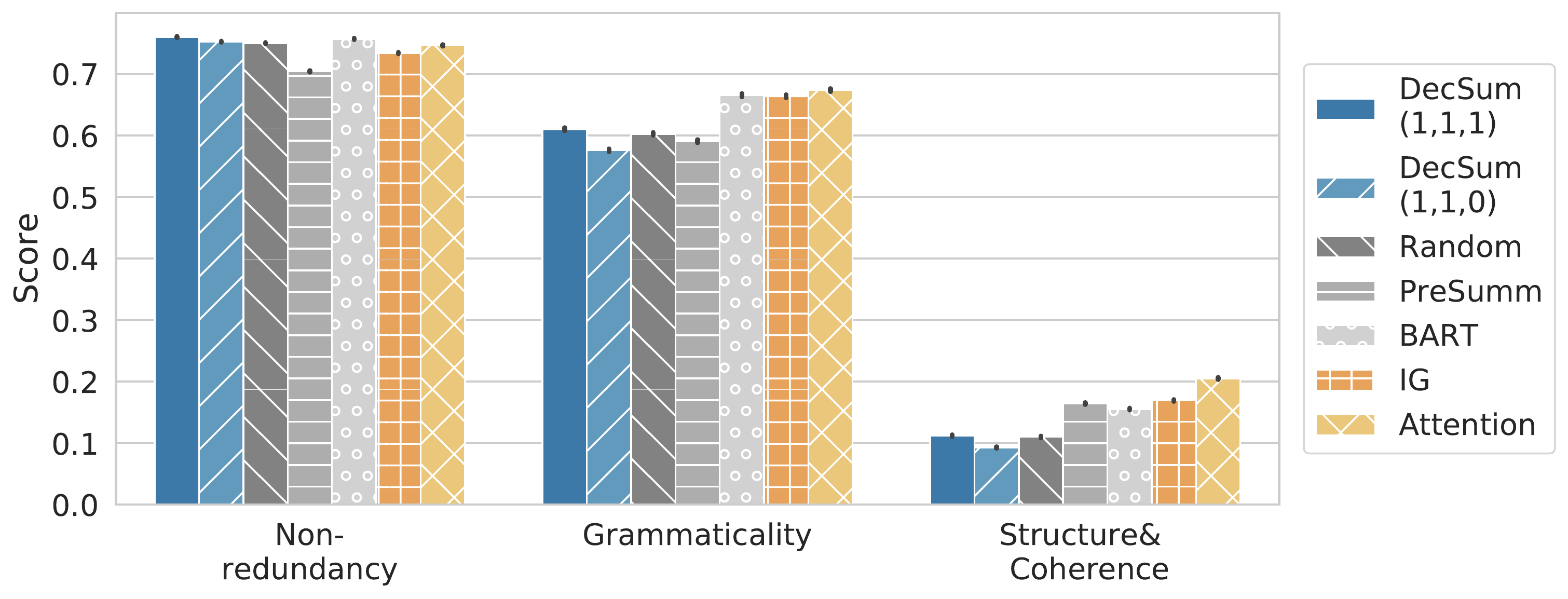}
    \caption{
    Summary quality evaluation using SUM-QE \cite{Xenouleas:EMNLP-IJCNLP19}.
    \system achieves strong \meaningdiversity, but leads to lower grammaticality and coherence.
    }
    \label{fig:quality}
\end{figure}

To further examine the effectiveness of our approach, we study the sentiment distribution using an independent classifier other than our own model.
We use a pretrained BERT model fine-tuned on sentiment analysis for product reivews\footnote{\url{https://huggingface.co/nlptown/bert-base-multilingual-uncased-sentiment}.} to determine the sentiment of sentences.
Specifically, the 5-class sentiment classification model outputs a class with the highest probability, and we define sentences with class 1 and 2 as negative, 3 as neutral, and 4 and 5 as positive.
Ideally, a representative summary should cover diverse sentiments.
\figref{fig:sentiment} shows that
\system can select a more diverse set of sentences with regard to sentiment diversity compared to other methods.
PreSumm and BART tend to select positive sentences over negative sentences which results in a less representative summary and can potentially mislead human decision making.
In comparison, model-based methods (i.e., IG and attention) tend to avoid neutral sentences.

\subsubsection{Text-only Summary Evaluation}
Finally, we evaluate \meaningdiversity and other text-only properties commonly used in standard text-only summarization (\figref{fig:quality}).
Overall, we find that \system achieves strong \meaningdiversity (0.760 vs. 0.757 with BART, $p=0.046$ with paired $t$-tests; comparisons with other baselines are all statistically significant with $p < 0.001$).
In comparison, PreSumm achieves the worst non-redundancy among the baselines.
Explanation methods (IG and attention) also provide worse non-redundancy than \system, as they do not explicitly optimize \meaningdiversity.

Meanwhile, \system leads 
to inferior performance based on other text-only evaluation metrics such as grammaticality and coherence.
Textual non-redundancy improves the grammaticality and coherence compared to (1, 1, 0).
Surprisingly, although attention does not take coherence into account, it leads to 
better coherence than text-only summarization methods.
We hypothesize that this is related to the fact that attention tends to select sentences that are more concentrated in sentiment distribution.

\begin{table*}[t]
    \centering
    \scriptsize
    \begin{tabular}{lp{0.41\textwidth}p{0.41\textwidth}@{\hspace{3pt}}r}
        \toprule
        Method & Restaurant 1: \textcolor{blue}{IHOP} & Restaurant 2: \textcolor{blue}{Tasty Kabob} (rated better after 50 reviews.) &  \#correct \\
 \midrule
 PreSumm &           \textcolor{orange}{$\tilde{x}_1$}: I had a pancake combo with New York cheese cake pancakes and they were delicious ! ! !. \textcolor{orange}{$\tilde{x}_2$}: This place was great \textcolor{orange}{$\tilde{x}_3$}: I got to eat breakfast and watch the football game !. \textcolor{orange}{$\tilde{x}_4$}: Finally a local IHOP , great service and always delicious breakfast. \textcolor{orange}{$\tilde{x}_5$}: Nice clean place. &  \textcolor{orange}{$\tilde{x}_1$}: Also they have the best Persian Ice Cream which is only one flavor .... \textcolor{orange}{$\tilde{x}_2$}: what is the flavor?? \textcolor{orange}{$\tilde{x}_3$}: ( its a secret , you will have to go there and find out ! ). \textcolor{orange}{$\tilde{x}_4$}: Tasty Kabob is a must see on any Hookah bar tour. \textcolor{orange}{$\tilde{x}_5$}: Tasty Kabob , while among the best Persian restaurants in Arizona , falls short of Famous Kabob in Sacramento and many Los Angeles joints. &       1/3 \\\midrule
 \system &                 \textcolor{orange}{$\tilde{x}_1$}: Love this place and they got big screen TV'S always playing football, great idea. \textcolor{orange}{$\tilde{x}_2$}: \ul{My soup came out cold, our server forgot our drinks, and they just microwaved it to warm it up and it literally over cooked everything in the soup}. \textcolor{orange}{$\tilde{x}_3$}: I had a pancake combo with New York cheese cake pancakes and they were delicious!!! &                                                                                                         \textcolor{orange}{$\tilde{x}_1$}: Regardless, both versions were moist and very appealing. \textcolor{orange}{$\tilde{x}_2$}: If you thought you didn't like Persian food, this place will definitely make you think again. \textcolor{orange}{$\tilde{x}_3$}: It was a generous portion for two, but I found myself munching on it just to pass the time until our lunches came, not because it was exceptionally well done. &       3/3 \\
 \midrule
        &  \multicolumn{1}{c}{\includegraphics[width=.7\columnwidth]{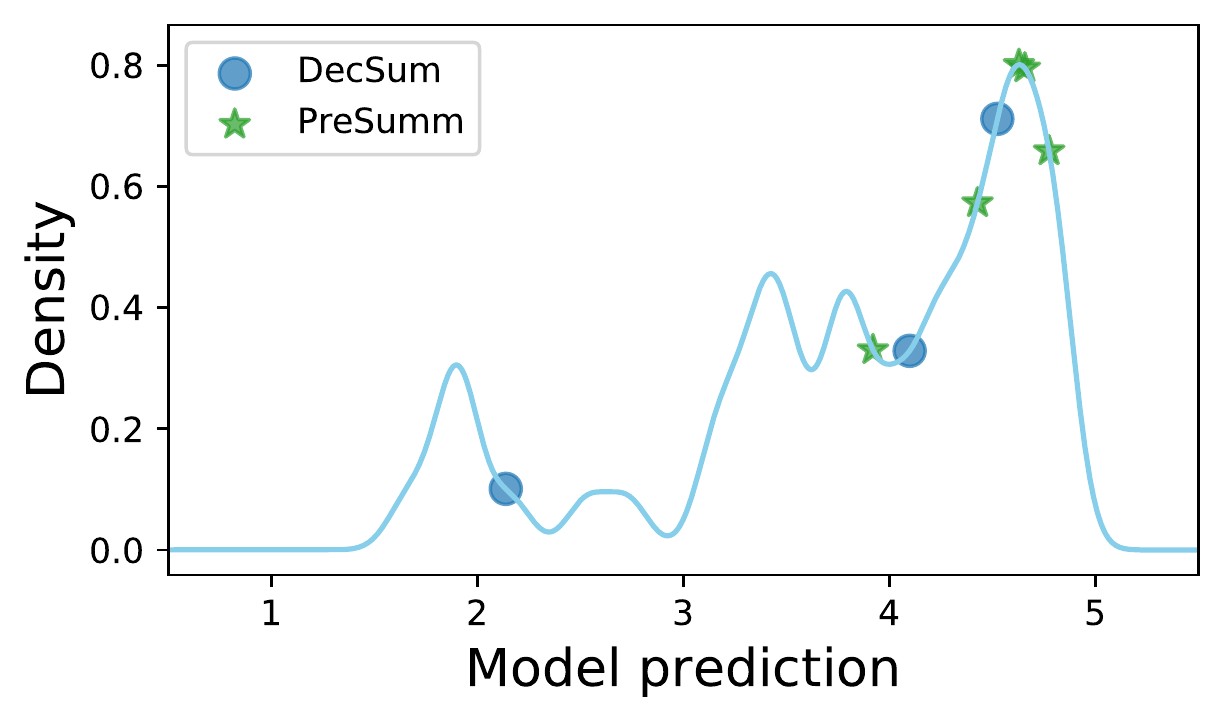}}
   & \multicolumn{1}{c}{\includegraphics[width=.7\columnwidth]{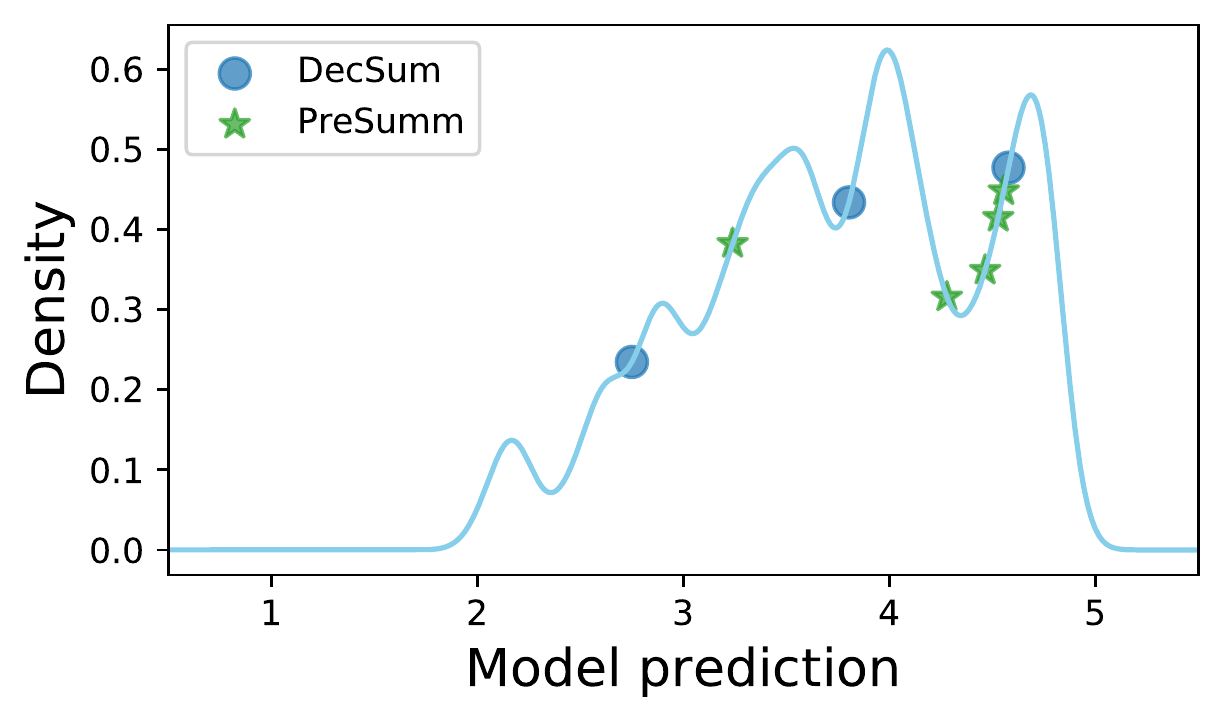}}&\\\bottomrule
    \end{tabular}
    \caption{Example summaries from PreSumm and \system  for two restaurants in Tempe, AZ. \textit{\#correct} is the result from human evaluation.
    Dots on the plots represent selected sentences on the distribution of model predictions.
    \system is able to capture sentence \textcolor{orange}{$\tilde{x}_2$} with a low predicted rating from reviews of IHOP to help participants distinguish future ratings between two restaurants.
    }
    \label{tab:summary}
\end{table*}

\subsection{Human Evaluation}
As the regression task is simplified to a binary classification task in human evaluations (\secref{sec:evaluation}), 
we first obtain model accuracy on the simplified task (\tableref{tab:model-human-exp}).
\system is the best summarization method with an accuracy of 76.1\%, comparable to using the full text.
Among our baselines, only PreSumm achieves above 60\% in the simplified task. 
We choose four methods for our human evaluation based on this result: random as a control condition, PreSumm as the better text-only summarization method, IG as our model-based explanation method, and \system.

\begin{table}[!t]
    \centering
    \small
    \begin{tabular}{lr>{\raggedleft\arraybackslash}p{2.5cm}}
        \toprule
        Method &      Accuracy (\%) &    Experiment subset accuracy (\%)  \\
        \midrule
        Full (oracle) & 76.0  & 85.5 \\\midrule
        \multicolumn{3}{c}{Text-only summarization methods} \\
        Random & 55.6 & 58.0 \\
        BART & 57.3 & 60.5 \\
        PreSumm & 64.4 & 66.0 \\ \midrule
        \multicolumn{3}{c}{Model-based explanation methods} \\
        IG & 57.3 & 59.0\\
        Attention & 52.8 & 52.5 \\ 
        \midrule
        \system 
        & 76.1 & 85.5\\
        \bottomrule
    \end{tabular}
    \caption{Model performance on the simplified binary classification task as described in \secref{sec:evaluation}. 
    We sample 200 restaurant pairs 
    for human evaluation.
    }
    \label{tab:model-human-exp}
\end{table}

\begin{figure}[t!]
    \centering
    \begin{subfigure}[t]{0.8\columnwidth}
        \includegraphics[width=1\columnwidth]{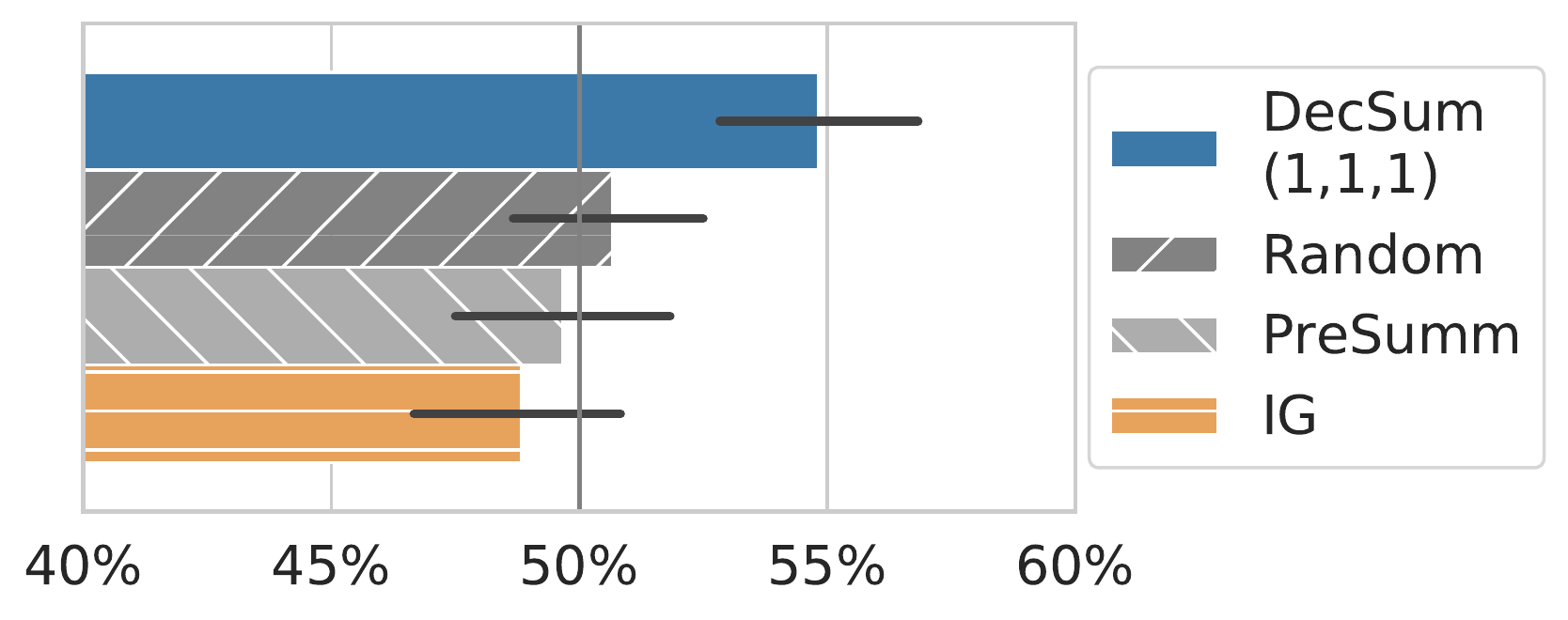}
        \caption{Human accuracy}
        \label{fig:human-acc}
    \end{subfigure}
    \begin{subfigure}[t]{\columnwidth}
        \includegraphics[width=1\columnwidth]{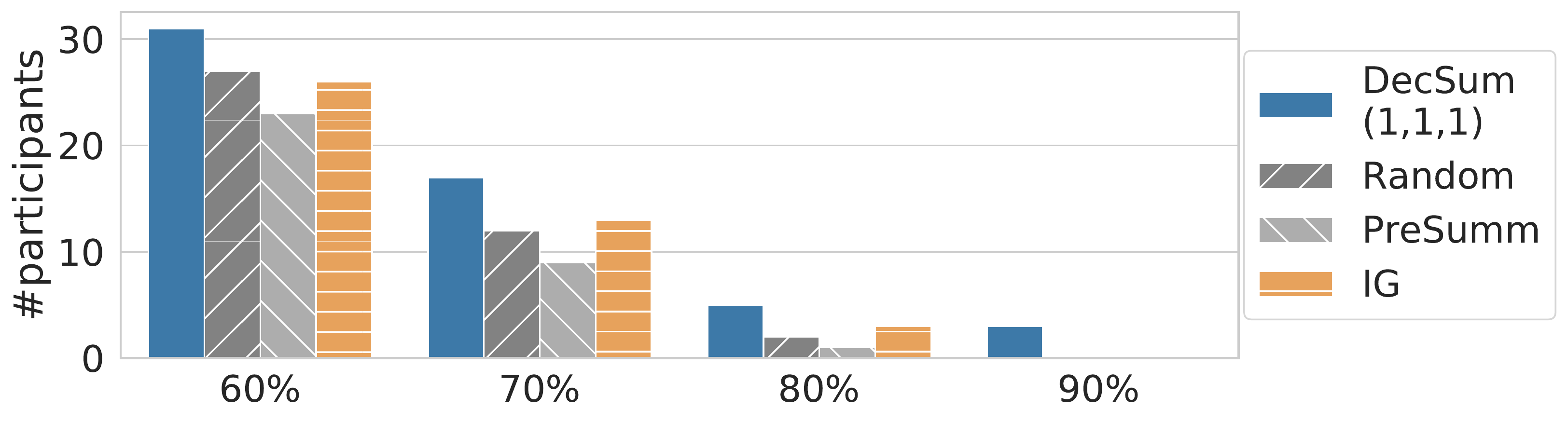}
        \caption{\#participants with over 60\% accuracy}
        \label{fig:human-dist}
    \end{subfigure}
    \caption{\figref{fig:human-acc} shows that
    \system is the only method that enables humans to statistically outperform random chance.
    \figref{fig:human-dist} further shows that \system leads to more individuals with high performance.
    }
\end{figure}

\figref{fig:human-acc} shows human performance in this simplified classification task.
This task turns out to be very challenging for humans and the best human performance is only 54.7\%, much lower than model accuracy in \tableref{tab:model-human-exp}.
This best human accuracy is achieved with \system and is statistically different from 50\% ($p=$0.017), while other baselines are all about chance (50.7\%, 49.7\%, and 48.8 for Random, PreSumm, and IG respectively; Random indeed slightly exceeds PreSumm and IG as it selects somewhat representative sentences).
\system also leads to more individuals with great performance:
three participants obtained 90\% accuracy with \system, but none with baseline methods did.
31 participants reached 60\%, 4 more than the second best (27 with random).

Although text-only qualities show that summaries from \system are less grammatical and coherent, the effect on human perception of usefulness is limited.
For instance, while 16 participants with IG strongly agree that summaries were useful in helping decide future ratings compared to 12 with \system, 15 with \system strongly agree that summaries were useful in helping assess confidence compared to 10 with IG.

Finally, \tableref{tab:summary} shows summaries of the same restaurant pair from \system and PreSumm, and 
the distribution plots present the corresponding selected sentences on the distribution of model predictions from all sentences.
Summaries from \system can better present the overall distribution and allow participants to evaluate these two restaurants.
For example, \system includes a negative sentence ($\tilde{x}_2$) from IHOP reviews to help users determine that IHOP is not better rated.
In contrast, PreSumm only selects positive sentences and fails to form a decision-representative summary.

\section{Related Work}

We review additional related work in three areas: query/aspect-based summarization, forecasting with NLP, and evaluation of summarization.

Our problem formulation is closely related to query-focused summarization \citep{daume-iii-marcu-2006-bayesian,wang-etal-2014-query,schilder2008fastsum,damova2010query}. 
In fact, \citet{wang-cardie-2012-focused} also uses the term ``decision'' and provides summaries for each decision made in a meeting.
Note that relevance in query-focused summarization is still based on textual information, whereas we incorporate potential insights about a decision from a supervised model into the summarization framework.
For example, query-focused summarization for pancreatic cancer may summarize all sentences that mention pancreas, but a supervised model may learn that smoking relates to pancreatic cancer and our approach then includes smoking history in the summary.

Similar to our work,
aspect-based summarization uses a predictive model to provide summaries for food, service, decor for reviews \citep{titov2008joint}.
Another related direction is identifying helpful sentences in product reviews
\cite{gamzu-etal-2021-identifying}.
It is useful to highlight our motivation in support decision making in {\em challenging} tasks 
towards effective human-AI collaboration \citep{green2019principles,lai+liu+tan:20,lai+tan:19}.
Unlike tasks such as textual entailment where models aim to emulate human intelligence, forecasting future outcomes, such as stock markets \citep{xing2018natural} and message popularity \citep{tan+lee+pang:14},
is challenging both for humans and for machines. 
Humans and machines may offer complementary insights in these tasks.
We chose restaurant rating prediction as an example about which laypeople may have valid intuitions.
We thus also propose novel desiderata, decision faithfulness and decision representativeness.

Evaluation of summarization is very challenging, partly because the goal of summarization is usually vague \citep{nenkova2004evaluating,fabbri2021summeval}.
Popular metrics such as ROUGE require reference summaries \citep{lin2004rouge}, but it is unclear that humans can provide useful summaries for decision making in challenging tasks given their limited performance and the scale of inputs.
Our formulation adopts a task-driven evaluation, i.e., human performance on the decision task which the summaries are supposed to support.
This resembles application-based evaluation of explanations in interpretability \citep{doshi2017towards}.

\section{Conclusion}

We propose a novel task, decision-focused summarization, and demonstrate that  \system outperforms text-only summarization methods and model-explanation methods in both automatic metrics and human evaluation.
There are many exciting future directions in advancing decision-focused summarization to support human decision making.
In particular, our human evaluation demonstrates a substantial gap between human performance and model performance.
One possible approach is to leverage visualizations similar to \figref{fig:dist} to enable interactive summarization so that users can see the decision variance and explore the textual information beyond a static set of sentences.
As humans are final decision makers in a wide variety of high-stake scenarios, ranging from healthcare to justice systems,
it is critical to investigate human-centered approaches to support human decision making.

\para{Ethics considerations.}
Our work promotes intelligent models that can be used to support human decision making.
We advocate the perspective of augmented intelligence: the goal of our system is to best support humans as final decision makers instead of maximizing model performance.
However, in decisions with fairness concerns (e.g., bailing decisions), important future directions include examining fairness-related metrics for the summaries and human-AI interaction.

\para{Acknowledgement.}
We thank anonymous reviewers for their valuable feedbacks. We thank Rebecca Willett, Kevin Gimpel, and members of the Chicago Human+AI Lab for their insightful suggestions.
This work is supported in part by research awards from Amazon, IBM, Salesforce, and NSF IIS-2125116, 2126602.

\bibliography{refs}
\bibliographystyle{acl_natbib}
 
\appendix
\section{Model Training Details and Comparisons with DAN and LR}
We fine-tune Longformer with 102M parameters on Nvidia RTX Titan GPU with half precision using Huggingface transformers package \cite{wolf-etal-2020-transformers}. 
We use AdamW \cite{loshchilov2017decoupled} optimizer with learning rate 5e-5 and linear warm-up of 500 steps.
We train Longormer for 3 epochs where the batch size is 4 and the maximum input token length is 3,000.
We search for hyperparameters for epochs $\{3, 4, 5\}$ and max sequence lengths $\{2000, 3000\}$ and choose the model checkpoint with lowest MSE on the validation set.
The model training time of Longformer is about an hour per epoch.
The beam search algorithm takes 3 days to find 15 sentences\footnote{We do not need that many sentences for the main paper, but we did that to understand the effect of summary length.} for processing the whole test set (3,623 restaurants) if the setting includes the faithfulness component.
Without faithfulness component, \system takes less than an hour on the test set.

\begin{figure*}[!t]
    \centering
    \begin{subfigure}[t]{0.32\textwidth}
        \centering
        \includegraphics[width=0.99\textwidth]{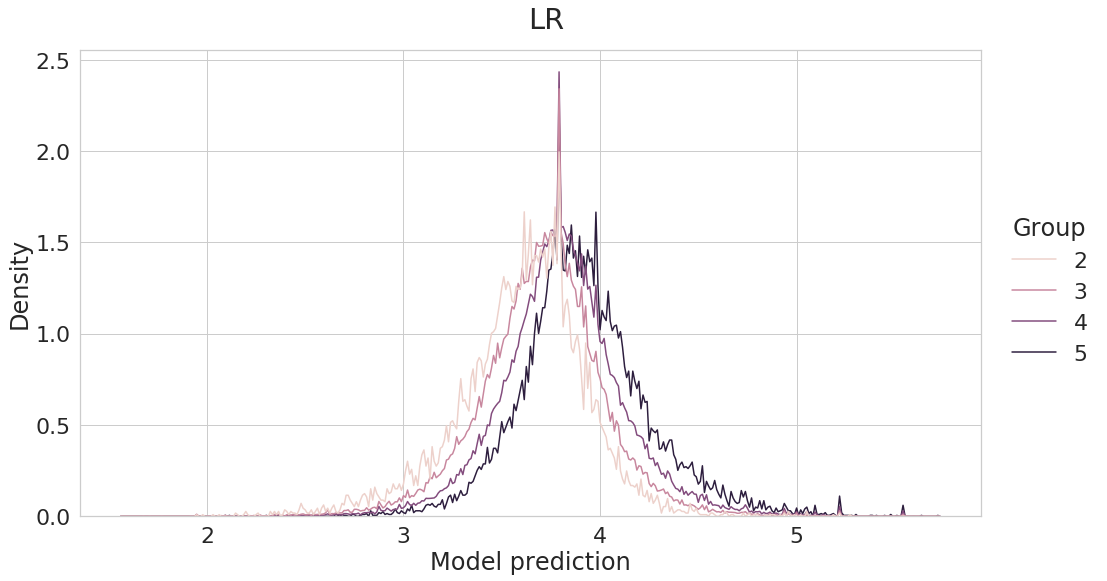}
        \caption{Logistic Regression.}
        \label{fig:lr-sent}
    \end{subfigure}%
    \begin{subfigure}[t]{0.32\textwidth}
        \centering
        \includegraphics[width=0.99\textwidth]{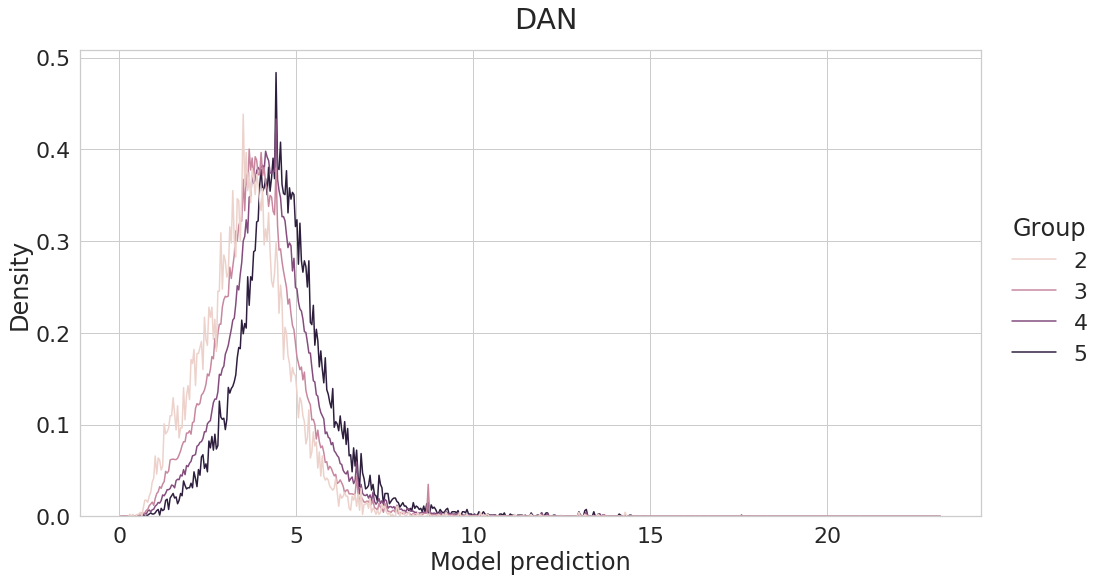}
        \caption{DAN.}
        \label{fig:dan-sent}
    \end{subfigure}%
    \begin{subfigure}[t]{0.32\textwidth}
        \centering
        \includegraphics[width=0.99\textwidth]{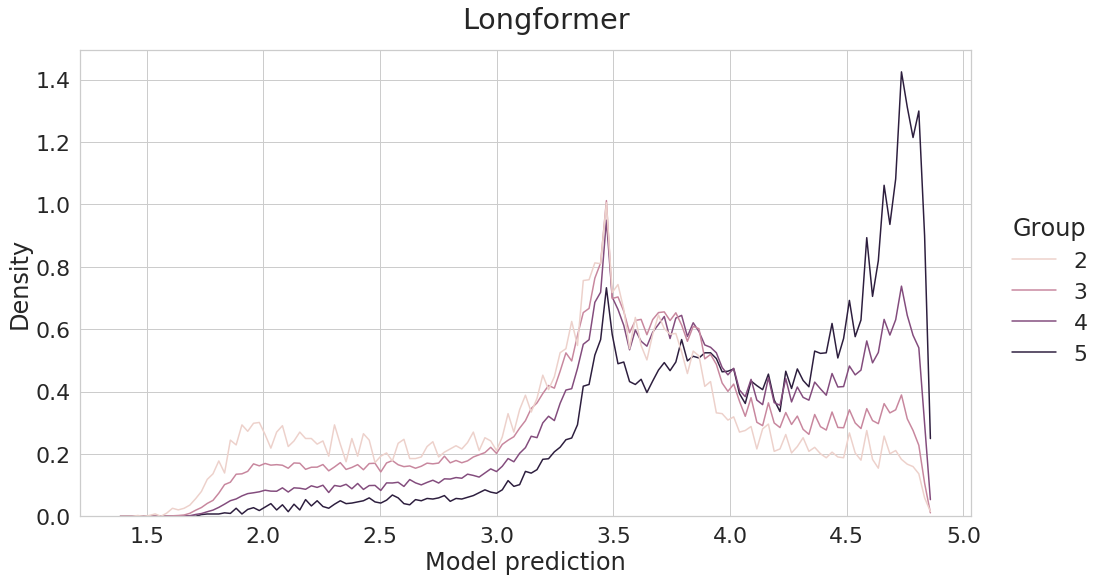}
        \caption{Longformer.}
        \label{fig:lr-sent}
    \end{subfigure}%

    \caption{Model prediction distributions of each rating group from 
    logistic regression (LR), 
    deep averaging networks (DAN), and Longformer. Only Longformer model can properly distinguish sentences located at different score range. LR and DAN are not robust to input length shift where models are trained with input of full 10 reviews but are tested with sentences.}
    \label{fig:sent-score}
\end{figure*}

Besides Longformer, we have tried logistic regression (LR) and Deep Averaging Networks (DAN) as our regression model. 
However, as shown in \figref{fig:sent-score}, only Longformer can provide appropriate prediction distributions of individual sentences.
We group restaurants into four groups where their average ratings of first 10 reviews are in $[1.5, 2.5)$, $[2.5, 3.5)$, $[3.5, 4.5)$, and $[4.5, 5]$ as group 2, 3, 4, and 5 respectively.
Then, we use a regression model trained with full 10 reviews $f: X \rightarrow y$ to predict ratings of individual sentences from different restaurants in the group.
Finally, we use Gaussian kernel density function to obtain the score distribution and plot sentence score distributions of different groups in the same figure.
Note that we do not show restaurants with ratings in the range of $[0, 1.5)$ because there are only a very small number of restaurants in this range.
We can see that the distributions from LR and DAN are close to normal distributions with different means for each group. 
More importantly, LR and DAN are not robust to distribution shift of input length, where the models are trained with full 10 reviews and are tested on individual sentences. LR can make predictions beyond 5 stars and DAN even makes predictions above 15.
In comparison, Longformer is able to distinguish positive, neural, and negative sentences and the distributions of different groups also reflect the sentiment distributions of each group. 

\begin{figure}[!t]
    \centering
    \begin{subfigure}[t]{0.9\columnwidth}
        \centering
   
        \includegraphics[width=\columnwidth]{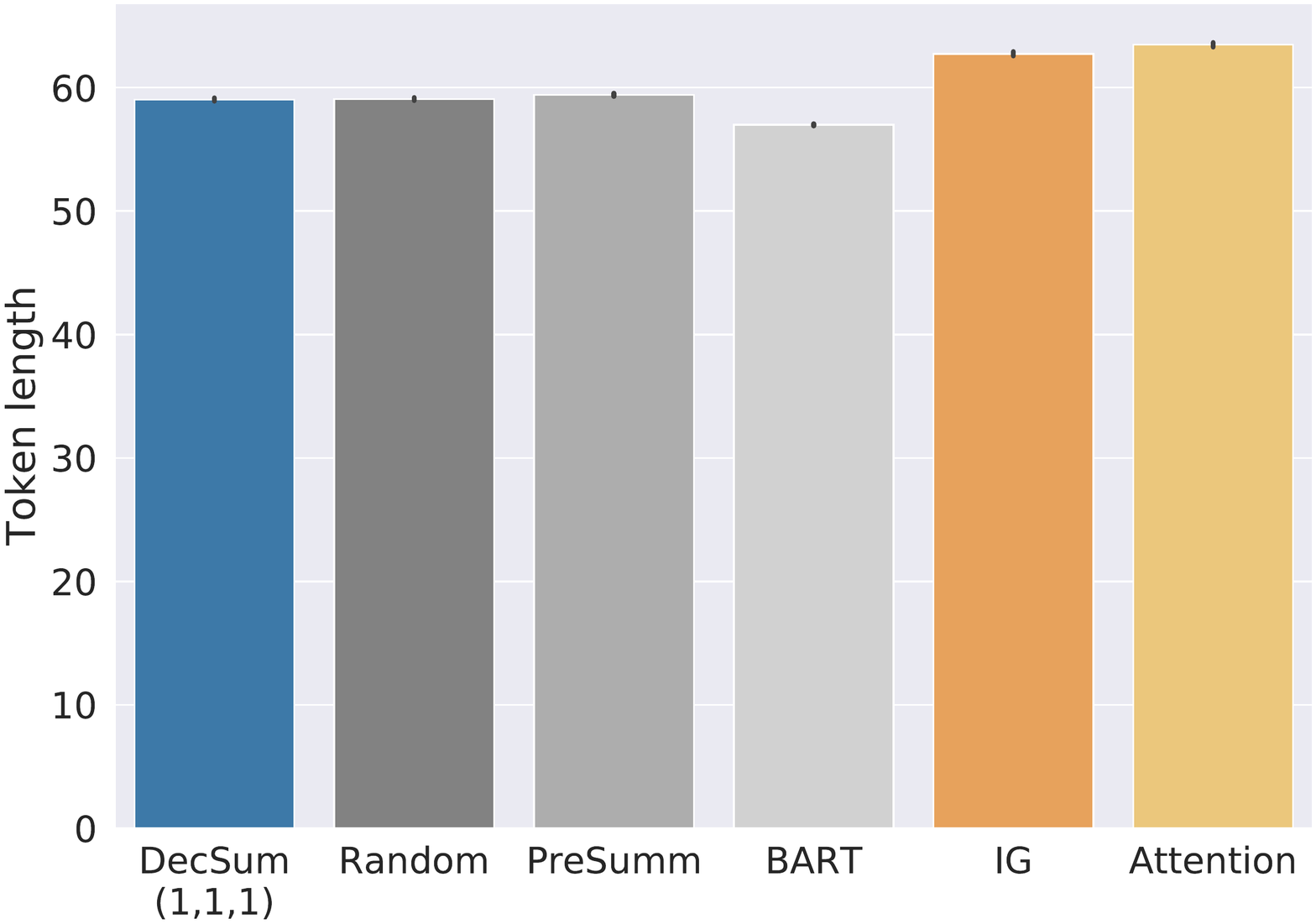}

        \caption{\small Token lengths of summarization approaches for human evaluation. The summary lengths are comparable after length truncation.
        }
        \label{fig:token-len}
    \end{subfigure}%

    \begin{subfigure}[t]{0.9\columnwidth}
        \centering
        \includegraphics[width=1\columnwidth]{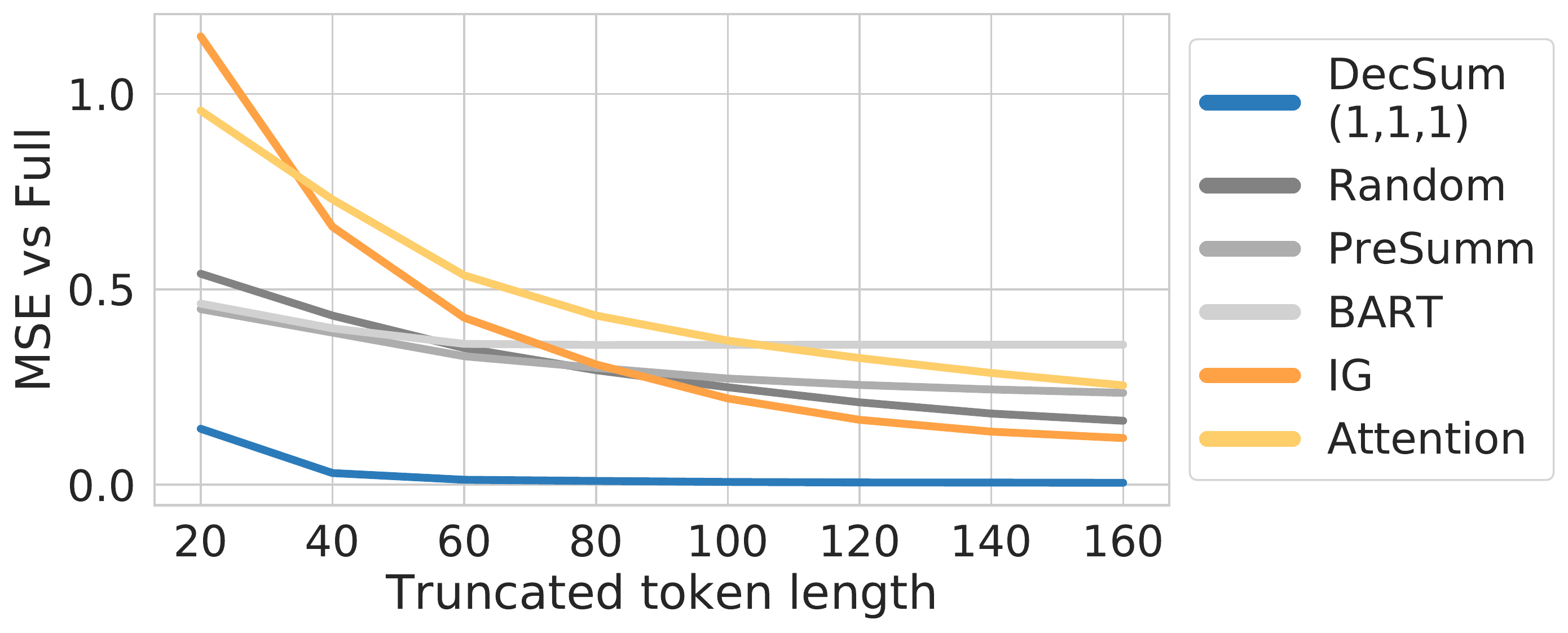}
        \caption{\small Faithfulness with predictions from all reviews using various token lengths. 
        }
        \label{fig:mse-all}
    \end{subfigure}%
 
    \begin{subfigure}[t]{0.9\columnwidth}
        \centering
   
        \includegraphics[width=1\columnwidth]{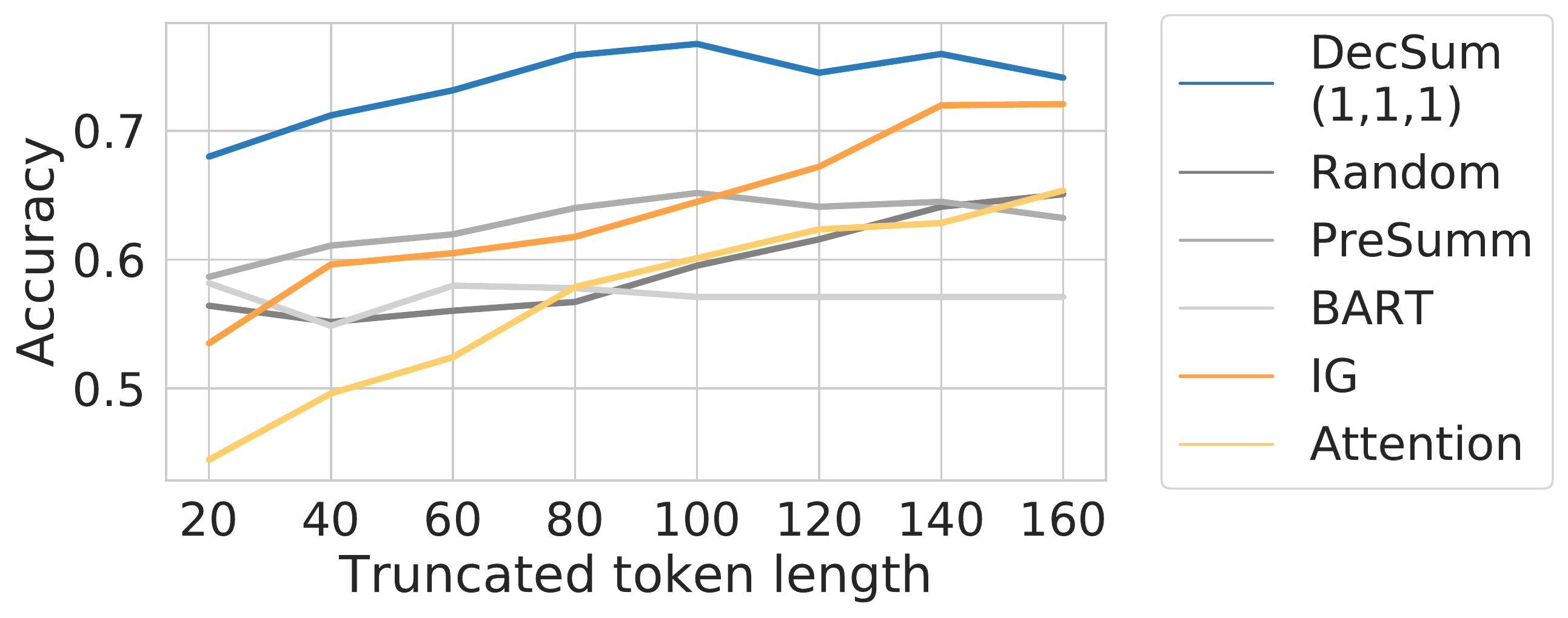}
        \caption{\small Model accuracy on the simplified classification task with different input token lengths.}
        \label{fig:length_effect}
    \end{subfigure}%

    \caption{The effect of summary length.}
    \label{fig:mse-len}
\end{figure}

\section{The Effect of Summary Length} 
To generate \system summaries in this paper, we use beam search to find 15 sentences for each restaurant and then truncate these sentences at the one that exceeds 50 tokens as summaries in our evaluation section.
\figref{fig:token-len} shows the average token length of different methods after controlling for the length. 
They are all comparable to each other.

Next, we investigate the effect of summary length in model prediction.
Note that we do hard truncation without considering the sentence boundaries in this section, so the results are not directly comparable to \tableref{tab:mse} and \tableref{tab:model-human-exp} in the main paper.
As show in \figref{fig:mse-len}, BART summaries do not improve along with the increase of token length because its average token length is only 60 where other extractive summarization approaches can select as many sentences as the full text in ten reviews. 
It's worth noting that random baseline becomes better than other baseline except IG after 100 tokens. 
The reason can be that random selection is more representative of the original reviews compared to PreSumm and attention methods. 
We also present model accuracy of the simplified task on various token lengths. 
\figref{fig:length_effect} shows \system still outperforms baselines substantially. 
PreSumm is the second best model but is surpassed by IG after 120 tokens.

\section{The Effect of Sentence Order} 
While computing the score of decision faithfulness component in \system algorithm, we concatenate the selected sentences in the original order of the first ten reviews.  
We find that the LongFormer supervised model is sensitive to the sentence order of summary. 
For example, for three selected sentences $x_1, x_4, x_8$ from the first ten reviews $X=\{x_1,x_2,...,x_i,...,x_N\}$ where $i$ is sentence index of concatenated first ten reviews, summary constructed from the selected order of \system, e.g., $x_8, x_1, x_4$, yields different results from summary with the original order $x_1, x_4, x_8$.
As shown in \tableref{tab:appendix-mse} and \tableref{tab:appendix-model-human-exp}, summaries built from selected order, which is different from \system algorithm, weaken the performance of \system on the decision faithfulness objective, and diminish the predictive power of the supervised model on simplified binary classification task.
Thus, building a supervised model which is robust to different sentence orders in the summary can be a future direction to pursue.

\begin{table}
    \centering
    \small
    \begin{tabular}{lrr}
        \toprule
        Method &  \shortstack{MSE with \textbf{Full} \\ (faithfulness) $\downarrow$}  &      MSE $\downarrow$ \\
        \midrule
        Full (oracle) & 0 & 0.135 \\
        \midrule
        \multicolumn{3}{p{0.9\linewidth}}{\system(selected order) ~ w/ ($\alpha$ decision faithfulness, $\beta$ decision representativeness, $\gamma$ \meaningdiversity)} \\ %
        (1, 1, 1)  & \underline{0.028}& 0.157 \\
        (1, 1, 0)  & \underline{0.076}& 0.200 \\ %
        (1, 0, 1) & \textbf{\underline{0.024}} & 0.154  \\ %
        (0, 1, 1)  & 0.174 & 0.288\\ %
        (1, 0, 0) & \underline{0.069} & 0.188 \\ 
        (0, 1, 0)  & 0.180 & 0.290\\ %
        (0, 0, 1)  & 0.537 & 0.588\\ %

        \bottomrule
    \end{tabular}
    \caption{
    MSE of model predictions based on summaries of \system where the sentences are concatenated with the \textbf{selected order} which is different from \system algorithm. 
    }
    \label{tab:appendix-mse}
\end{table}

\begin{table}[!t]
    \centering
    \small
    \begin{tabular}{lr>{\raggedleft\arraybackslash}p{2.5cm}}
        \toprule
        Method &      Accuracy (\%) &    Experiment subset accuracy (\%)  \\
        \midrule
        Full (oracle) & 76.0  & 85.5 \\
        \midrule
        \system \\
        (original order) & 76.1 & 85.5\\
        (selected order) & 73.8 & 75.0\\
        \bottomrule
    \end{tabular}
    \caption{Comparison between \system with different sentence order methods on the simplified binary classification task.
    }
    \label{tab:appendix-model-human-exp}
\end{table}

\section{Human Evaluation Details and Additional Results}
To choose 200 restaurant pairs for human evaluation, we randomly select from eligible restaurant pairs and limit restaurants per city to 25.
We make sure a restaurant does not appear twice in a HIT with 10 restaurant pairs.
In the end, 320 restaurants are used in human study, including 2 restaurants for the example pair.
In human evaluation, we disallow duplicate participants in our HITs by checking the worker id. 
We rejected 5 assignments for submitting a confirmation code but not actually doing the experiment.
The human study takes about 10 minutes for crowdworkers on average.

In the exit survey, many participants found our experiment interesting and the experience was smooth. They also shared the heuristics used while doing the HITs.
For example, \textit{``For the most part, I considered the tone of the reviews. If one review had a more positive tone than the other, I figured that one would get better reviews in the future''} and \textit{``I only used the summaries. I decided based on what I thought seemed like it was an ongoing issue. I didn't read too much into them if it seemed like it was a one-off issue.''} Some people may rely on information beyond reviews: \textit{``I focused mostly on the summaries. However, when summaries weren't enough I also focused on the locations and names.''}
As for the experiment experience, one participant indicated, \textit{``I really enjoyed this survey, and it was unique/different in many aspects, and one of my favorite things to do is read reviews so it was actually fun for me.''}.
Another crowdworker said, \textit{``The experiment was easy to follow and enjoyable because it was not like any others.''}
Also, \textit{``I basically felt like I was guessing considering I got the practice question wrong but I did give my earnest best answers. Interesting and engaging, thank you.''}
However, a small fraction of participants found that the 20-second timer is too long and preferred a timer of 10 or 15 seconds.

\begin{figure}[t]
    \centering
    \begin{subfigure}[t]{0.23\textwidth}
        \centering
        \includegraphics[width=1\textwidth]{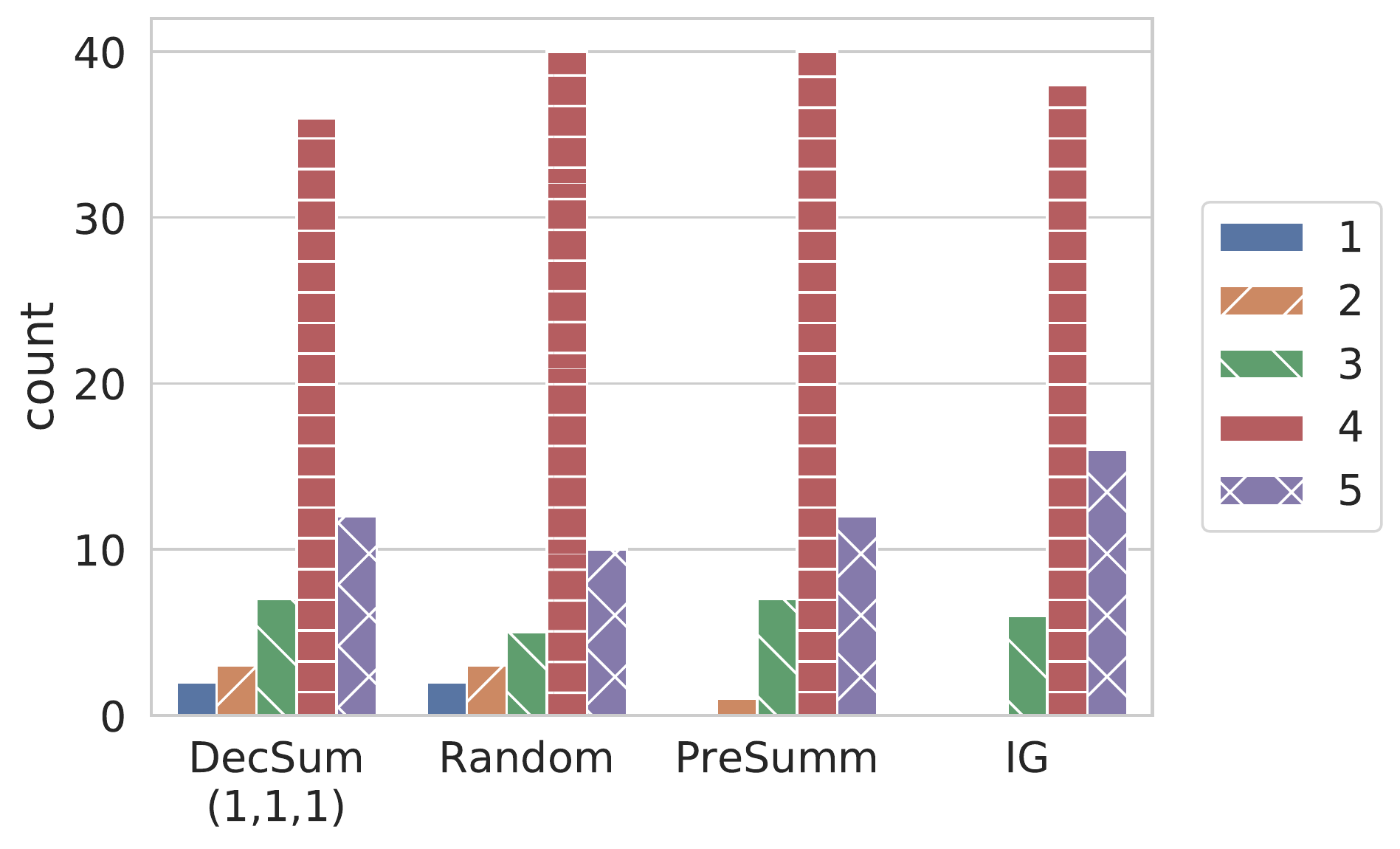}
        \caption{Usefulness in deciding rating.}
        \label{fig:usefull_decision}
    \end{subfigure}
    \begin{subfigure}[t]{0.23\textwidth}
        \centering
        \includegraphics[width=1\textwidth]{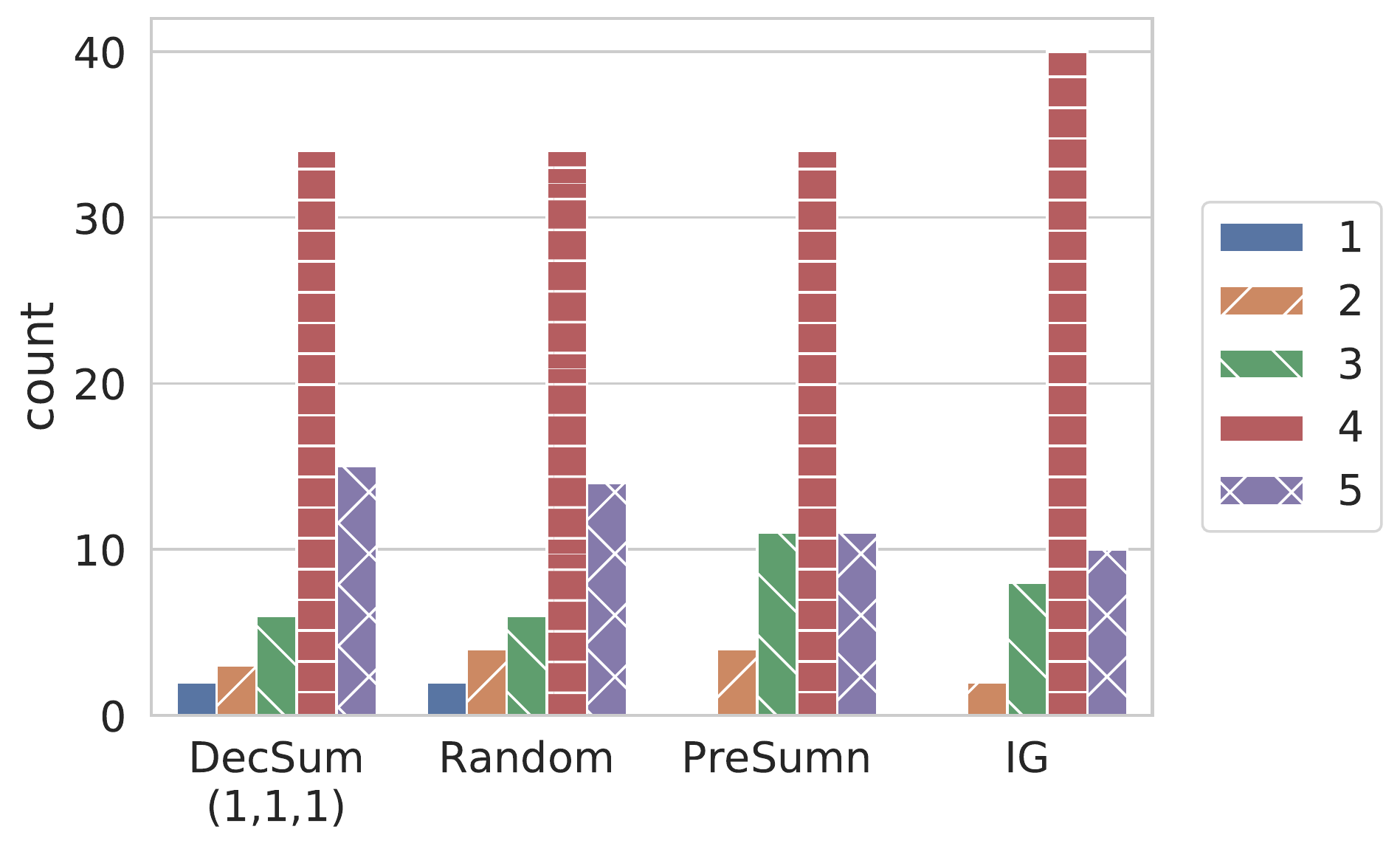}
        \caption{Usefulness in assessing confidence.}
        \label{fig:usefull_confidence}
    \end{subfigure}
    \caption{Self-reported usefulness.}
    \label{fig:usefulness}
\end{figure}

Participants provided self-reported usefulness rating as shown in \figref{fig:usefulness}. In general, these self evaluations are not correlated to the actual performance on the simplified task.

\begin{figure}[!t]
        \includegraphics[width=\columnwidth]{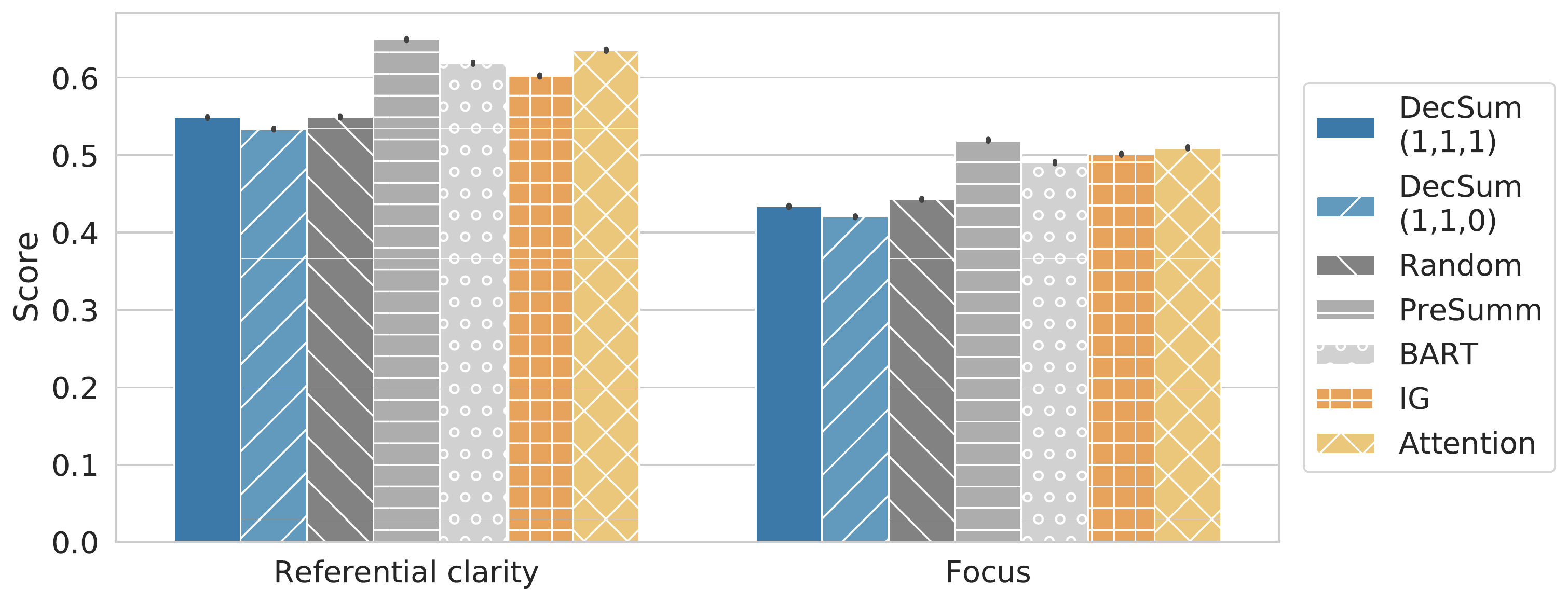}
    \caption{SUM-QE evluation on referential clarity and focus.
    }
    \label{fig:quality-appendix}
\end{figure}

\section{Summary Quality}
\figref{fig:quality-appendix} shows two additional summary quality evaluations with SUM-QE, referential clarity and focus. As \system encourages textual non-redundancy, \system is worse than text-only summarizations and model-based explanations on these two metrics.

\section{More Example Summaries from Experiment Subset}
In \tableref{tab:example1} and \tableref{tab:example2}, we present more example restaurant pairs along with model prediction distributions. 
   \begin{table*}[h!]\centering\small\begin{tabular}{p{0.07\linewidth} | p{0.43\linewidth} | p{0.43\linewidth}}
   \toprule

           method &                                                                                                                                                                                                                                                                                                                                                                                                                                                                                                                                                                    Jimmy John's &                                                                                                                                                                                                                                                                                                                                                                                                                                                                        Bueno Burger (rated better after 50 reviews.)\\
           \midrule
           Random &                             \textcolor{orange}{$\tilde{x}_1$}: I was impressed - not only was it fresh, but the bread was delicious. \textcolor{orange}{$\tilde{x}_2$}: All of the kids acted completely un interested in making sure we had what we needed. \textcolor{orange}{$\tilde{x}_3$}: My decision:  a turkey unwich. \textcolor{orange}{$\tilde{x}_4$}: The only thing that I was disappointed with \textcolor{orange}{$\tilde{x}_5$}: Now I crave sandwiches from there. &                                                                  \textcolor{orange}{$\tilde{x}_1$}: They cook the patties to pink in the middle, so if you like \textcolor{orange}{$\tilde{x}_2$}: Not to mention the flavor from the mesquite grill which is unlike anything else out there. \textcolor{orange}{$\tilde{x}_3$}: Went here with a couple of other people. \textcolor{orange}{$\tilde{x}_4$}: Perhaps they didn't finish constructing the place? \textcolor{orange}{$\tilde{x}_5$}: My friend who ordered the hot dog said the bun was hard, and he felt it was stale vs. being toasted.\\
           \midrule PreSumm &  \textcolor{orange}{$\tilde{x}_1$}: I highly recommend JJ 's on Hayden ! ! ! !. \textcolor{orange}{$\tilde{x}_2$}: I noticed that a new Jimmy John 's had opened on Hayden in McCormick Ranch. \textcolor{orange}{$\tilde{x}_3$}: We used to order sandwiches a couple times a week at work in Seattle. \textcolor{orange}{$\tilde{x}_4$}: When I moved to Scottsdale about a year ago , I would make the drive to North Scottsdale ( 8 - 10 miles ) just for a delicious sandwich. &  \textcolor{orange}{$\tilde{x}_1$}: I split the an Arizona style and a Glendale style burger. \textcolor{orange}{$\tilde{x}_2$}: This restaurant just opened 10/21/11. \textcolor{orange}{$\tilde{x}_3$}: Gary also brought around a " Arizona " style burger that had been mis - ordered. \textcolor{orange}{$\tilde{x}_4$}: The Arizona burger was definitely the better of the two. \textcolor{orange}{$\tilde{x}_5$}: Gary hooked my wife up with their super - fire - burner - hot sauce , and while I ca n't do those types of things , my wife said it has great flavor and was indeed very hot.  \\
           \midrule IG &                                 \textcolor{orange}{$\tilde{x}_1$}: Great food, amazing customer service, and a great atmosphere. \textcolor{orange}{$\tilde{x}_2$}: Terrific bread, great tasting sandwich! Music was too loud to hold a conversation and the staff seem disinterested. \textcolor{orange}{$\tilde{x}_3$}: Food was great but employees were disgusting. \textcolor{orange}{$\tilde{x}_4$}: Was told they couldn't and the reason is "it's against company policy". &                                                                                                                                                                                                        \textcolor{orange}{$\tilde{x}_1$}: With so many burger joints out there offering a lot of the same, Bueno Burger offers fresh local ingredients and a unique menu which allows you to customize your burger experience. \textcolor{orange}{$\tilde{x}_2$}: The tables are rickety, the lighting is weird, and particle board design has not sufficiently replaced the skeleton of Boston Market. \\
       \midrule \system &                       \textcolor{orange}{$\tilde{x}_1$}: Thank you, Jimmy John's!  :) \textcolor{orange}{$\tilde{x}_2$}: It took everything inside of me not to walk back in and put them in their place. \textcolor{orange}{$\tilde{x}_3$}: Thank you Jimmy John's, for adding a little brightness to my day. \textcolor{orange}{$\tilde{x}_4$}: Freaky fast! \textcolor{orange}{$\tilde{x}_5$}: The kids pointed and laughed behind his back while mocking him as he walked away. &                                                                                                                                                                                                            \textcolor{orange}{$\tilde{x}_1$}: The chimi and burger were full sized, however I'm used to a bit more fries (and/or rings), but personally i'm trying to avoid the "super size" mentality, so it's fine by me. \textcolor{orange}{$\tilde{x}_2$}: Showing up at a new restaurant on opening day is a real treat because usually it's about the only time you'll see owners and managers.   \\                  \midrule&                                                                                                                                                                                                                                                                                                                                                                                                                                                                                                                                \includegraphics[width=\linewidth]{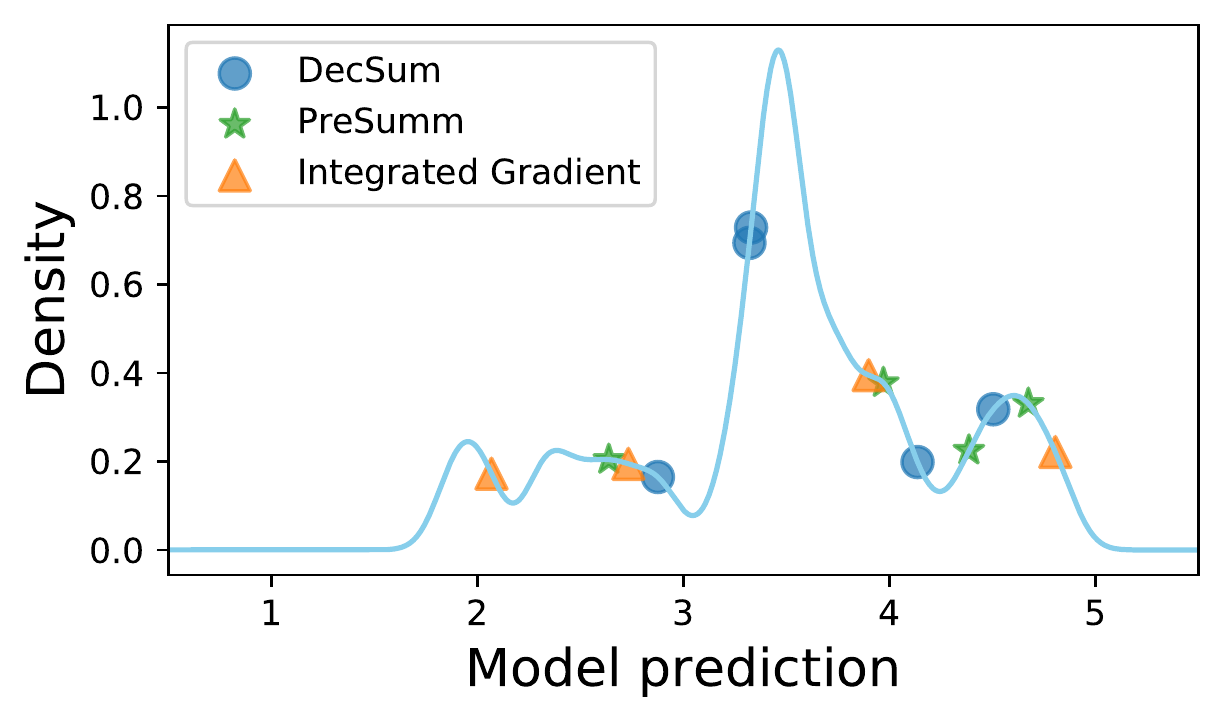} &                                                                                                                                                                                                                                                                                                                                                                                                                         \includegraphics[width=\linewidth]{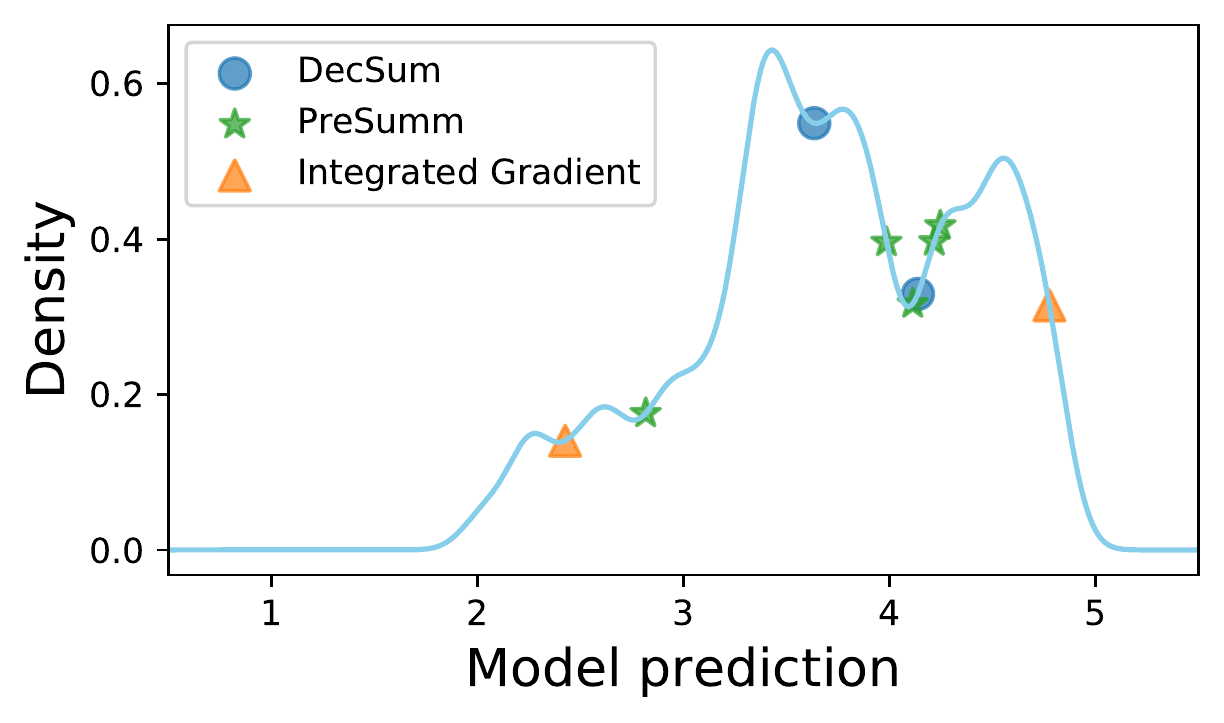} \\
   \bottomrule
   
   \end{tabular}
   \caption{\label{tab:example1}  Restaurant pair at Scottsdale, AZ.}
   \end{table*}

\begin{table*}[h!]\centering\small\begin{tabular}{p{0.07\linewidth} | p{0.43\linewidth} | p{0.43\linewidth}}
    \toprule
      
     method &                                                                                                                                                                                                                                                                                                                                                                                                                                                       Village Pub \& Poker (rated better after 50 reviews.) &                                                                                                                                                                                                                                                                                                                                                                                                                                      Cafe Pan \\
     \midrule     Random &                                                                                       \textcolor{orange}{$\tilde{x}_1$}: but I sure was glad to have ordered my burger \textcolor{orange}{$\tilde{x}_2$}: he told me" the girls". \textcolor{orange}{$\tilde{x}_3$}: the prime rib was OK... \textcolor{orange}{$\tilde{x}_4$}: They started going to the Village Pub several months ago and have been telling me how great the food is and how reasonable the prices were, so I was looking forward to it. &  \textcolor{orange}{$\tilde{x}_1$}: They are in the same location as Cocolini but have changed the name. \textcolor{orange}{$\tilde{x}_2$}: They are in the same location as Cocolini but have changed the name. \textcolor{orange}{$\tilde{x}_3$}: and it was wonderful! \textcolor{orange}{$\tilde{x}_4$}: Not too bad in the new Vegas. \textcolor{orange}{$\tilde{x}_5$}: Arrangement is a bit slapped together for a \$12 dessert crepe. \\
     \midrule  PreSumm &  \textcolor{orange}{$\tilde{x}_1$}: great food good prices \textcolor{orange}{$\tilde{x}_2$}: 1.00 ellis island beer \textcolor{orange}{$\tilde{x}_3$}: 6.99 steak salad bake potatoe as good as its  gets \textcolor{orange}{$\tilde{x}_4$}: i felt guilty when i paid my bill \textcolor{orange}{$\tilde{x}_5$}: i thought they  made a mistake \textcolor{orange}{$\tilde{x}_6$}: This is our favorite place. \textcolor{orange}{$\tilde{x}_7$}: This is a local chain w/ locations all over the valley. &                                                          \textcolor{orange}{$\tilde{x}_1$}: The ham and cheese croissant sandwich was a great on - the - go breakfast. \textcolor{orange}{$\tilde{x}_2$}: Located in the food court off the casino floor of the Venetian. \textcolor{orange}{$\tilde{x}_3$}: Not too bad in the new Vegas. \textcolor{orange}{$\tilde{x}_4$}: Located in the food court off the casino floor of the Venetian. \\
     \midrule     IG &                                                                                                                                                                                         \textcolor{orange}{$\tilde{x}_1$}: They started going to the Village Pub several months ago and have been telling me how great the food is and how reasonable the prices were, so I was looking forward to it. \textcolor{orange}{$\tilde{x}_2$}: This was by far the worst service I have received in a long time. &                                         \textcolor{orange}{$\tilde{x}_1$}: Had a very bad experience here. \textcolor{orange}{$\tilde{x}_2$}: It was our first time eating at this place and we definetly wouldn't recommend it to anybody else. \textcolor{orange}{$\tilde{x}_3$}: I went back later for gelato, and that was incredible, as well. \textcolor{orange}{$\tilde{x}_4$}: Possibly the best espresso I've had outside of Europe. \\
     \midrule \system &                                                                                                                                                \textcolor{orange}{$\tilde{x}_1$}: Oh, how I wish that this place was able to take advantage of its Desert Shores location and offer outside dining on the lake, but it's angled location makes that impossible.   \textcolor{orange}{$\tilde{x}_2$}: The food was okay and the prices were reasonable, but unless my parents are treating I'm not going back. &            \textcolor{orange}{$\tilde{x}_1$}: It is near other food places, almost like a food court and plenty of seating available.   \textcolor{orange}{$\tilde{x}_2$}: Will update this review next time after I try them. \textcolor{orange}{$\tilde{x}_3$}: but they're known for their crepes, gelato and what I usually get is the waffles. \textcolor{orange}{$\tilde{x}_4$}: There are MANY restaurants and coffee shops to eat at...  \\
     \midrule &                                                                                                                                                                                                                                                                                                                                                                                                                                                  \includegraphics[width=\linewidth]{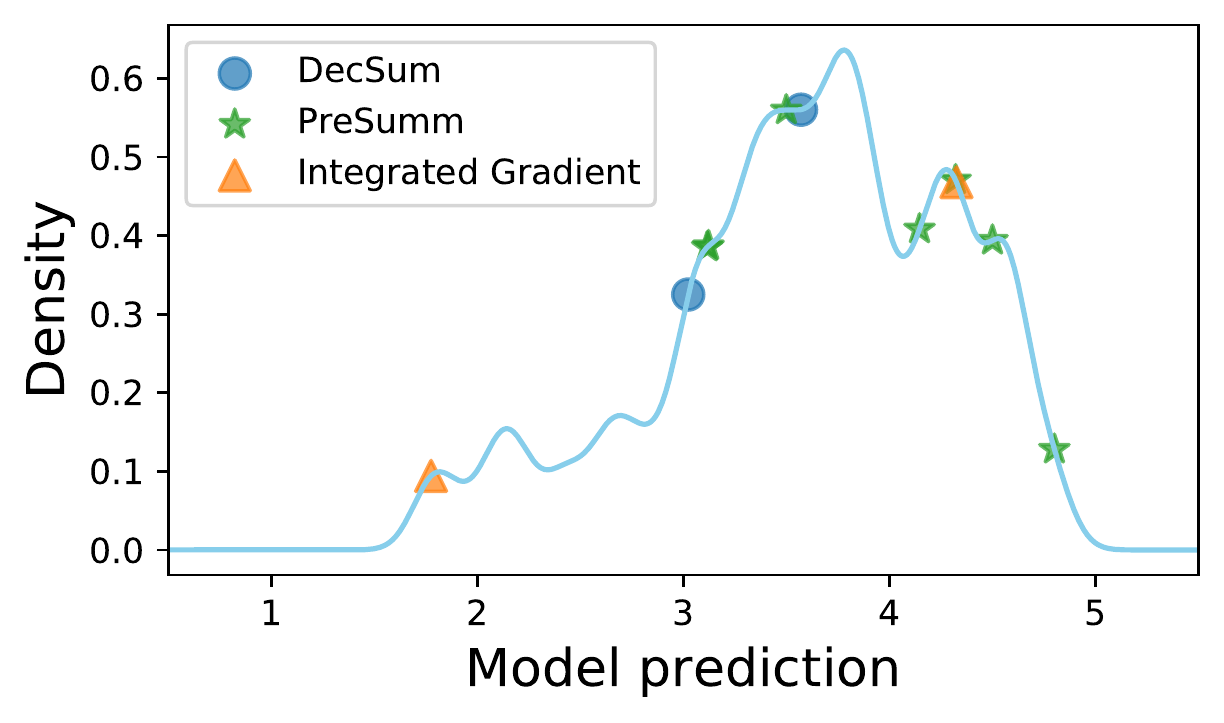} &                                                                                                                                                                                                                                                                                                                                                                                    \includegraphics[width=\linewidth]{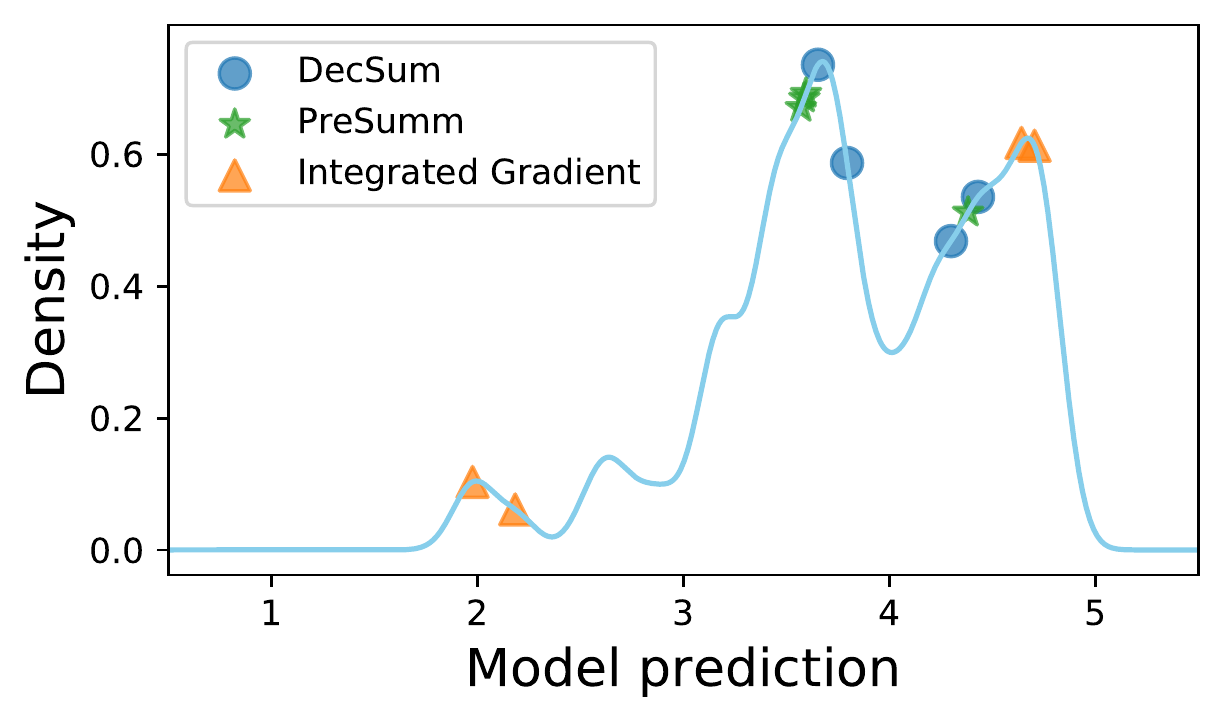}  \\
\bottomrule
\end{tabular}
\caption{\label{tab:example2} Restaurant pair at Las Vegas, NV.}
\end{table*}

\end{document}